%% file: main.tex
\documentclass[]{sensenova}

\definecolor{sensepurple}{HTML}{1565C0}
\usepackage{graphicx}
\input{mmstyles.sty}
\usepackage{wrapfig}
\usepackage{booktabs}
\usepackage{fontawesome5}
\usepackage{colortbl}
\usepackage{arydshln} % for dashed lines \hdashline

\definecolor{rowblue}{RGB}{220, 230, 242}
\definecolor{deltaup}{RGB}{0, 140, 0}
\definecolor{deltadn}{RGB}{200, 0, 0}

\newcommand{\up}[1]{{\scriptsize\textcolor{deltaup}{+#1}}}
\newcommand{\dn}[1]{{\scriptsize\textcolor{deltadn}{-#1}}}

\newcommand{\chairdn}[1]{{\scriptsize\textcolor{deltaup}{-#1}}}
\newcommand{\chairup}[1]{{\scriptsize\textcolor{deltadn}{+#1}}}

\usepackage{multirow}
\usepackage[table]{xcolor}
\usepackage[framemethod=TikZ]{mdframed}

\usepackage{orcidlink}
\usepackage{hyperref}

\title{
    \begin{minipage}{0.12\textwidth}
        \raggedleft
        \includegraphics[height=1.9cm]{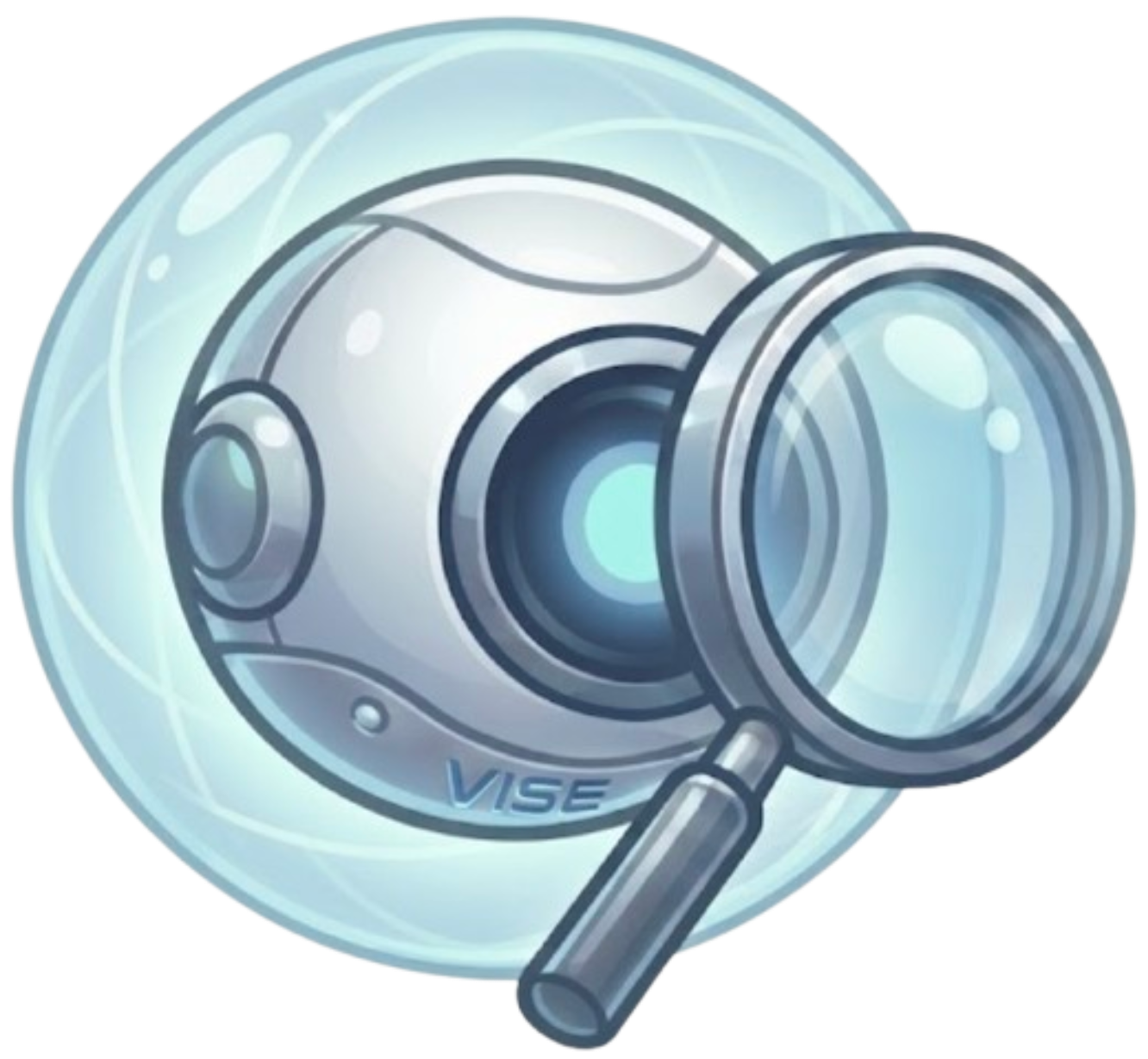}
    \end{minipage}%
    \hspace{-0.5cm}
    \begin{minipage}{0.8\textwidth}
        \centering
        \textbf{Paying More Attention to Visual Tokens in Self-Evolving Large Multimodal Models}
    \end{minipage}
}

\author{%
  \normalfont
  \textbf{Shravan Venkatraman}$^{1}$, \textbf{Ritesh Thawkar}$^{1}$, \textbf{Omkar Thawakar}$^{1}$, \textbf{Rao Muhammad Anwer}$^{1,2}$,\\
  \textbf{Hisham Cholakkal}$^{1}$, \textbf{Salman Khan}$^{1,3}$, \textbf{Fahad Shahbaz Khan}$^{1,4}$ \\[2pt]
  $^{1}$Mohamed bin Zayed University of Artificial Intelligence \quad
  $^{2}$Aalto University \\
  $^{3}$Australian National University \quad
  $^{4}$Linköping University
}

\input{sec/0_abstract}

\begin{document}
\maketitle

\clearpage

\input{sec/1_intro}
\input{sec/2_related_works}
\input{sec/3_model}
\input{sec/4_results}

\input{sec/5_conclusion}

\bibliographystyle{plainnat}
\bibliography{main}

\clearpage

\input{supplementary}

\end{document}

%% file: sec/0_abstract.tex
\abstract{

Recently, self-evolving large multimodal models (LMMs) have received 
attention for improving visual reasoning in a purely unsupervised 
setting. However, multi-role self-play and self-consistency reward 
schemes in existing self-evolving LMMs optimize answer agreement 
without ensuring the decoder attends to visual content, relying instead 
on statistical language priors to produce self-consistent outputs. This 
leads to a persistent failure mode we term visual under-conditioning, 
where the decoder relies on language priors rather than the image during 
generation, manifesting as insufficient attention to visual tokens. As a 
result, current self-evolving LMMs struggle on vision--language 
understanding tasks such as image captioning and visual question 
answering. To address this, we propose \textbf{VISE} (Visual Invariance 
Self-Evolution), a purely unsupervised self-evolving framework that 
directly regularizes the model's visual conditioning policy through two 
complementary invariance-based rewards: a geometric invariance reward 
that enforces spatial consistency under known transformations, and a 
semantic invariance reward that penalizes evidence-agnostic generation by 
requiring the model to recognize the absence of evidence when predicted 
regions are perturbed. VISE operates within a single model without 
specialist roles, external reward models, or annotations, and is trained 
on raw unlabeled images. Experiments on 18 benchmarks demonstrate the 
efficacy of our approach. Using Qwen3-VL-2B as the base model, VISE 
achieves gains of $+16.85$ CIDEr on COCO and $+19.66$ CIDEr on 
TextCaps, reduces object hallucination by $5.0$ Chair-I points, and 
generalizes across multiple model families and scales.

\begin{center}
\vspace{0.4em}
\includegraphics[width=\textwidth]{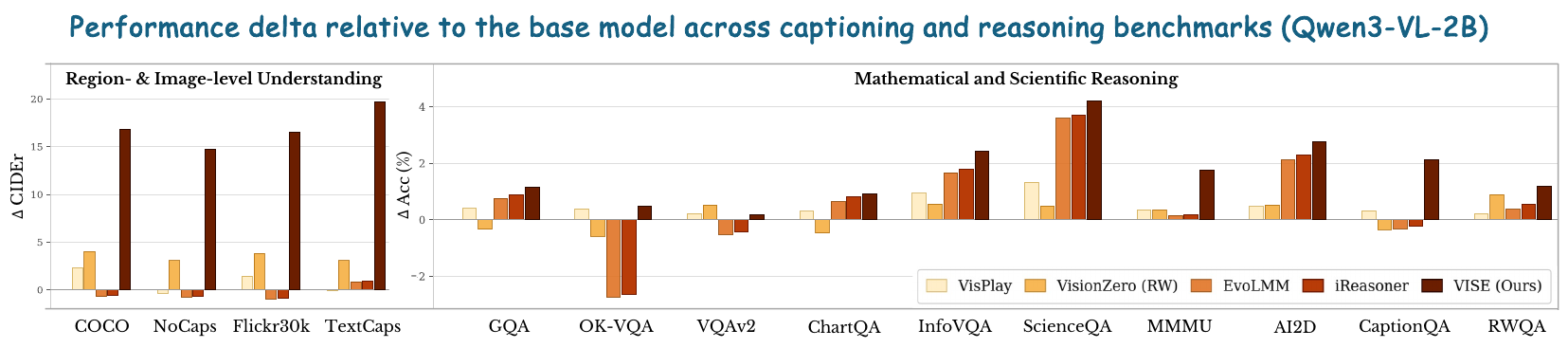}
\vspace{-1.2em}
\label{fig:comparison}
\end{center}

\checkdata[
\raisebox{-1pt}{\includegraphics[height=10pt]{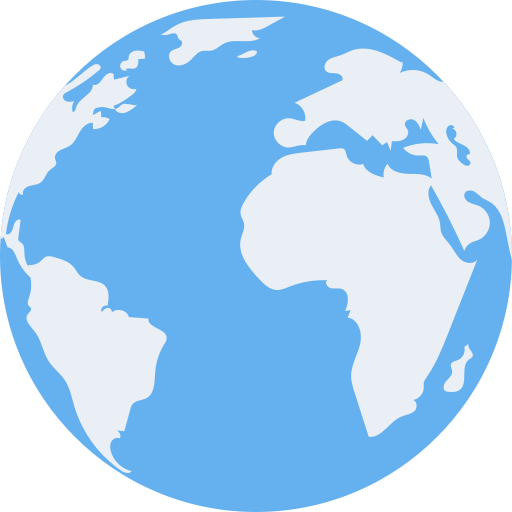}}\hspace{3pt}Project Page
]{\href{https://mbzuai-oryx.github.io/VISE/}{\texttt{mbzuai-oryx.github.io/VISE/}}}

\checkdata[
\raisebox{-2pt}{\includegraphics[height=10pt]{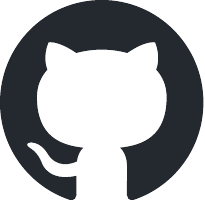}}\hspace{3pt}GitHub Code
]{\href{https://github.com/mbzuai-oryx/VISE}{\texttt{github.com/mbzuai-oryx/VISE}}}

\checkdata[
\raisebox{-2pt}{\includegraphics[height=10pt]{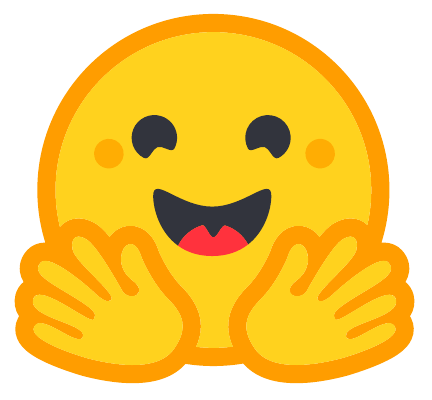}}\hspace{3pt}HuggingFace Model
]{\href{https://huggingface.co/shravvvv/VISE}{\texttt{huggingface.co/shravvvv/VISE}}}

}

%% file: sec/1_intro.tex
\section{Introduction}

% \vspace{-10pt}

Recent work on self-evolving large multimodal models (LMMs) has 
demonstrated that models can improve their visual reasoning capabilities 
in a purely unsupervised manner, without relying on human-annotated 
supervision or externally verified reward 
models~\cite{p3,p7,p1,visionzero}. These approaches frame multimodal 
reasoning as a self-improving process, instantiating proposer--solver or 
questioner--reasoner roles within a shared backbone and reinforcing 
multi-sample agreement, trajectory-aware feedback, and tool-assisted 
verification to promote more reliable reasoning~\cite{p1,p3,p6,p7}. 
Despite their promise, however, these methods optimize for answer 
agreement without ensuring the decoder attends to the actual visual 
content, relying instead on statistical language priors to produce 
self-consistent outputs. This leads to a persistent failure mode we term 
\textit{visual under-conditioning}, which causes hallucination, modality 
bypass, and unstable visual interpretation even when the vision encoder 
produces accurate representations. Concretely, this manifests as 
insufficient attention to visual tokens during decoding, where generations are 
driven more by language priors than by the image being described.

This limitation arises from two structural features common to existing
self-evolving frameworks. First, the Proposer--Solver setup forms an
implicit minimax game: the roles are optimized jointly with opposing
objectives, making training unstable over long horizons. In practice,
one role often dominates; Proposer collapses to trivial or
degenerate queries that guarantee agreement, or the Solver overfits to
the Proposer’s distribution and fails to generalize, driving the system
into local minima that are difficult to correct without external
intervention. Second, framing the reward around answer correctness
implicitly assumes that self-consistency implies correct visual
grounding. A decoder exploiting language priors can achieve high answer
agreement through statistical co-occurrences alone, leaving visual
under-conditioning unaddressed or even reinforced.

\begin{figure}[t]
    \centering
    \includegraphics[width=0.97\linewidth]{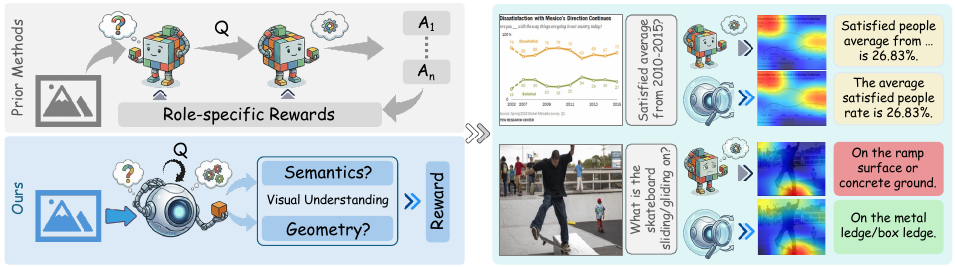}
    % \vspace{-10pt}
    \caption{\textbf{Overview of VISE compared to prior self-evolving methods.}~\cite{p7,p3,p1,visionzero} \textit{(Left)} Prior approaches use 
separate roles with role-specific objectives optimized for answer 
consistency, whereas our VISE operates within a single model using geometric 
and semantic invariance rewards. \textit{(Right)} For chart-based queries that require minimal visual dependence, both the prior method and our VISE provide accurate answers. However, on real-scene understanding tasks that require deeper visual semantics, prior approaches struggle, likely because the language decoder resorts to statistically plausible scenarios (a ``ramp surface'' instead of a ``metal ledge''). In contrast, VISE accurately identifies the metal ledge, likely due to learning self-consistent visual invariance during unsupervised training. Additional examples are provided in the suppl. material.}

    \label{fig:teaser}
% \vspace{-20pt}
\end{figure}

As illustrated in Fig.~\ref{fig:teaser}, this failure mode is directly 
observable. On chart-based or structured scientific queries, prior self-evolving 
methods produce the same correct answers as ours, suggesting that the required 
visual cues may be sparse or quickly resolved, after which the decoder can 
proceed largely from its internal representations. However, on natural scene understanding queries such as identifying what a skateboard is actually resting 
on, these methods default to statistically common associations (ramp 
surface, concrete ground), indicating that visual evidence is not being 
robustly integrated throughout the generation process. In contrast, correctly resolving such cases requires the decoder to 
remain conditioned on the specific image content, exposing the gap 
between answer consistency and genuine visual interpretation. 
Prior self-evolving methods trained on science- and math-centric images
show gains on reasoning benchmarks, yet perform at or below the base
model on captioning and region-level description tasks that require
detailed visual conditioning. This disparity validates the hypothesis that improvements in
answer agreement do not translate into stronger visual interpretation,
and that visual under-conditioning persists in existing self-evolving
frameworks.

% \begin{figure}[t]
%     \centering
%     \includegraphics[width=\linewidth]{assets/Plots.pdf}
%     % \vspace{-20pt}
%     \caption{\textbf{Performance delta relative to the base model across 
% captioning and reasoning benchmarks (Qwen3-VL-2B).} Delta CIDEr 
% (left) and delta accuracy (right) for five self-evolving methods. 
% Methods trained on math- and science-centric images improve structured 
% reasoning (e.g., EvoLMM $+3.59$ on ScienceQA) but regress on 
% captioning, whereas VISE achieves the largest captioning gains 
% ($+16.85$ COCO, $+19.66$ TextCaps) while remaining competitive on 
% reasoning tasks.}
%     \label{fig:comparison}
% % \vspace{-17pt}
% \end{figure}

To directly address this, we propose \textbf{VISE} (\textbf{V}isual
\textbf{I}nvariance \textbf{S}elf-\textbf{E}volution), a purely
unsupervised self-evolving framework that regularizes the model's
visual conditioning policy rather than answer agreement. Unlike
prior multi-role formulations, VISE operates within a single model,
motivated by the observation that a well-pretrained LMM already
possesses sufficient knowledge to formulate meaningful queries about
its visual content. Its training signal comes from two
complementary invariance-based rewards: a \textit{geometric invariance}
reward that enforces spatial consistency under known transformations,
and a \textit{semantic invariance} reward that requires the model to
recognize the absence of evidence when predicted regions are
perturbed. Training on raw images with no annotations, metadata,
or external reward models, VISE achieves gains of up to $+16.85$
CIDEr on COCO and $+19.66$ CIDEr on TextCaps, reduces object
hallucination (Chair-I) by $5.0$ points, and improves consistently
across VQA and reasoning benchmarks, demonstrating that stronger
visual conditioning generalizes across tasks rather than trading off
against them. Mechanistically, VISE achieves this by increasing attention to visual 
tokens across decoder layers during generation, reflecting the 
shift from language-prior-driven to image-conditioned decoding.

In summary, our main contributions are as follows:
% \vspace{-5pt}
\begin{itemize}
    \item We introduce VISE, a single-model, fully unsupervised self-evolving 
    framework that directly addresses visual under-conditioning in LMMs by 
    regularizing the model's visual conditioning policy rather than its answer 
    agreement, requiring no annotations, metadata, or external reward models.
    \item We propose two complementary invariance-based reward signals: geometric 
    invariance and semantic invariance via regional perturbation (ghosting), that 
    jointly enforce spatial consistency and evidence sensitivity, providing a 
    purely self-supervised objective defined entirely from the model's own 
    predictions.
    \item We empirically validate VISE across 18 benchmarks spanning image 
    captioning, VQA, reasoning, and hallucination on multiple model scales and 
    four backbone families, demonstrating consistent and substantial improvements 
    over all self-evolving baselines with no tradeoffs between task groups.
\end{itemize}

% \vspace{-15pt}

%% file: sec/2_related_works.tex
\section{Related Work}
\label{sec:relatedwork}

% \vspace{-11pt}

% \noindent\textbf{Self-Evolution in Large Multimodal Models.}
Early work on self-improving LMMs aimed to reduce reliance on human-annotated data through internal preference and alignment signals. Tan~\etal~\cite{beyond} introduce image-driven self-questioning with diffusion-based rejection for DPO optimization. Wang~\etal~\cite{sima} propose an in-context self-critic with visual metrics to construct preference pairs without external models. Liu~\etal~\cite{p10} study multimodal self-evolution from a reinforcement learning perspective and introduce entropy-driven exploration to mitigate saturation. While these approaches reduce annotation dependence, they still rely on structured supervision or curated preference construction, limiting fully autonomous self-improvement.

More recent work has shifted to fully unsupervised self-play
frameworks with internally generated rewards, yielding strong gains
on structured reasoning benchmarks. EvoLMM~\cite{p3} introduces a
Proposer--Solver formulation optimized with continuous
self-consistency rewards to enable co-evolution of question generation
and reasoning without annotations. Subsequent extensions refine this
paradigm: iReasoner~\cite{p1} incorporates trajectory-aware rewards,
VisPlay~\cite{p7} promotes diversity and difficulty to prevent
collapse, and Agent0-VL~\cite{p6} integrates tool-grounded
self-verification. C2-Evo~\cite{p4} and DoGe~\cite{p8} further
address training instabilities through co-evolutionary data loops and
role decoupling mechanisms.

\noindent\textbf{Limitations.}
Despite their methodological diversity and strong performance on
structured reasoning benchmarks, these approaches optimize answer
correctness or reasoning consistency as the primary objective. This
implicitly assumes that self-consistent outputs reflect improved
visual understanding, which is an assumption that breaks down under visual
under-conditioning, where a model can remain self-consistent and even
correct while relying on statistical language priors rather than
visual evidence. VISE addresses this directly by replacing
answer-agreement rewards with invariance-based rewards that
regularize the model's visual-conditioning policy itself, operating within a
single model on raw unlabeled images without specialist roles or
external reward models.

%% file: sec/3_model.tex
\begin{figure}[t]
    \centering
    \includegraphics[width=0.8\linewidth]{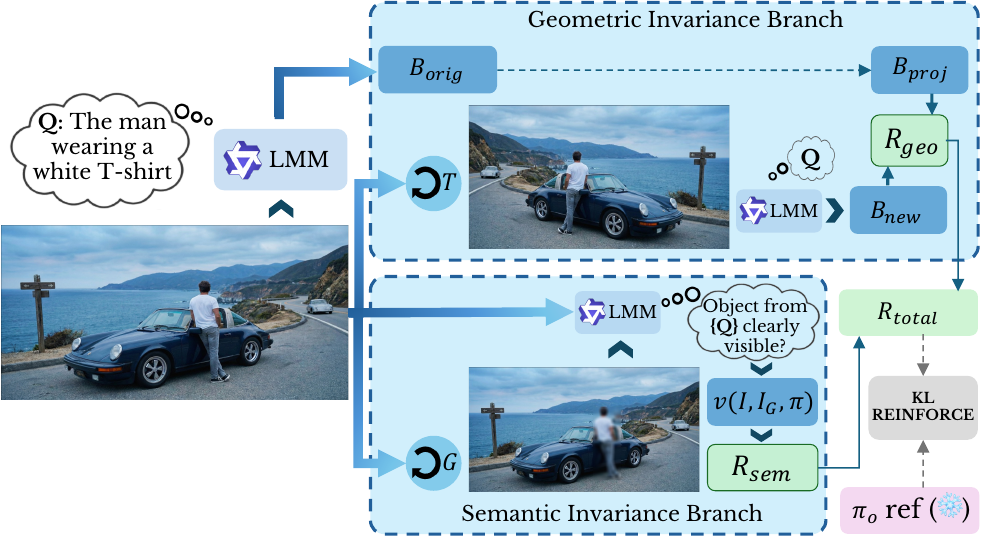}
    % \vspace{-9pt}
    \caption{\textbf{Overview of the VISE self-evolving framework.} Given a raw
unlabeled image, the model first generates a localization query and predicts a
bounding box $B_\text{orig}$. The Geometric Invariance Branch applies a spatial transformation $\mathcal{T}$, predicts $B_\text{new}$ on
the transformed view, and computes $\mathcal{R}_\text{geo}$ as the GIoU
between $B_\text{new}$ and the projected box $B_\text{proj}$ to enforce
spatial consistency across views. The Semantic Invariance Branch ghosts
the predicted region via blurring and assigns
$\mathcal{R}_\text{sem}$ only if the model detects the object before
perturbation and not afterward, penalizing evidence-agnostic generation.
The combined reward is optimized with KL-regularized REINFORCE against
a frozen reference policy $\pi_o$, without annotations, external reward
models, or specialist roles.}
\label{fig:architecture}
% \vspace{-19pt}
\end{figure}

% \vspace{-4pt}

\section{Method}

% \vspace{-11pt}

% Large multimodal models (LMMs) frequently fail not because their visual
% features are inadequate, but because the language decoder learns to
% bypass them. Given sufficiently strong language priors, the decoder
% can achieve high-likelihood outputs by relying on statistical
% co-occurrences from pre-training rather than attending to the actual
% image content---a failure mode we term \textit{visual under-conditioning}.
% The result is systematic shortcutting, hallucination, and unstable
% region-level grounding, even when the underlying vision encoder
% produces rich and accurate representations. Existing self-evolving
% frameworks address related problems through multi-role co-evolution
% (proposer--solver, challenger--judge), but they optimize for answer
% correctness rather than visual interpretation consistency, and
% therefore do not directly target the decoder shortcutting problem.
% We propose \textbf{VISE} (\textbf{V}isual \textbf{I}nvariance
% \textbf{S}elf-\textbf{E}volution), a fully unsupervised self-evolving
% framework that strengthens the model's \textit{evidence binding}---the
% degree to which its generated text and localization decisions
% consistently depend on the observed image---by optimizing two
% invariance-based reward signals derived entirely from the model's
% own predictions.

\noindent\textbf{Problem Formulation.}
Let $\mathcal{X} = \{x\}$ denote an unlabeled collection of images,
with no accompanying queries, bounding box annotations, or category
labels. At each training step, the model $\pi_\theta$ operates in a
self-questioning regime: it first generates a natural-language
localization query $q$ by interrogating image $x$, then predicts a
bounding box $B = (x_1, y_1, x_2, y_2)$ locating the queried object
under the same query. Both query generation and bounding box prediction are performed by the 
same single policy, without separate specialist roles, as a 
well-pretrained LMM already possesses sufficient visual knowledge to 
formulate meaningful queries for its own content. All spatial
coordinates are represented in a normalized space $[0, S]^4$, where
$S = 1000$, such that a pixel-space coordinate $c_\text{pix}$ along
dimension $D$ maps to $\tilde{c} = (c_\text{pix} / D) \cdot S$.

We use the Qwen3-VL family \cite{bai2025qwen3vltechnicalreport} as the base backbone, freezing the vision encoder and updating the multimodal projector, feed-forward layers, and decoder attention projections. This is motivated by the nature of the failure mode: the 
vision encoder already produces strong visual representations, and 
the problem lies in how the decoder projects and utilizes them. 
Propagating noisy, unsupervised reward gradients into the encoder 
would risk destabilizing representations that are already high 
quality, without addressing the locus of failure.

% \vspace{-2pt}

\noindent\textbf{Geometric Invariance Reward.}
Visual under-conditioning manifests most directly in localization
instability: a decoder that disengages with visual image evidence will
produce predictions that are inconsistent with the actual spatial
structure of the scene, and in particular will fail to maintain
coherent localization when the image undergoes a known geometric
transformation. We exploit this as a self-supervised training signal.
If the model correctly conditions its localization on what it sees,
then its predicted box on a geometrically transformed image should
correspond precisely to the analytically projected version of its
prediction on the original. Deviation from this consistency is
evidence of visual under-conditioning, and we penalize it directly.

At each training step, the model generates a query $q$ by
conditioning on image $x$, producing a short natural-language
description of a prominent and spatially unambiguous object in the
scene. The model then predicts a bounding box $B_\text{orig}$ on the
image under query $q$. A geometric 
transformation $\mathcal{T}$ is sampled uniformly from 
three types: affine (rotation $\theta \sim \mathcal{U}(-10^\circ, 
10^\circ)$, scale $s \sim \mathcal{U}(0.9, 1.1)$, translation 
$(\delta_x, \delta_y) \sim \mathcal{U}(-50, 50)^2$), crop (ratio 
$\rho \sim \mathcal{U}(0.8, 1.0)$, resized to original resolution), 
or horizontal flip. Each is described by a known $3{\times}3$ 
homogeneous matrix $M$. The model then predicts a second box
$B_\text{new}$ on the
transformed image $x' = \mathcal{T}(x)$ under the same query $q$.
The expected box under the transformation is computed
by lifting the four corners of $B_\text{orig}$ to homogeneous
coordinates and applying $M$ as $\mathbf{c}'_i = M\mathbf{c}_i$;
the axis-aligned box of the resulting corners gives the
projected box $B_\text{proj}$. The geometric invariance reward is:

% \vspace{-9pt}

\begin{equation}
    \mathcal{R}_\text{geo} =
    \frac{\,\text{GIoU}(B_\text{proj},\, B_\text{new}) + 1\,}{2}
    \label{eq:r_geo}
\end{equation}

% \vspace{-5pt}

\noindent where GIoU is the Generalized Intersection over Union~\cite{giou}:

% \vspace{-5pt}

\begin{equation}
    \text{GIoU}(B_1, B_2) = \text{IoU}(B_1, B_2) -
    \frac{|\mathcal{C}| - |B_1 \cup B_2|}{|\mathcal{C}|}
    \label{eq:giou}
\end{equation}

% \vspace{-3pt}

\noindent with $\mathcal{C}$ denoting the smallest axis-aligned box
enclosing both $B_1$ and $B_2$. The linear normalization in
Eq.~(\ref{eq:r_geo}) maps GIoU $\in [-1, 1]$ to
$\mathcal{R}_\text{geo} \in [0, 1]$. This reward is maximized
when the model's localization on the transformed view agrees
precisely with the geometric projection of its original prediction,
and degrades smoothly as spatial consistency deteriorates. In this
way, $\mathcal{R}_\text{geo}$ directly targets the spatial dimension
of visual under-conditioning: such a model cannot maintain such consistency
across views and is therefore penalized, even if its individual
predictions appear plausible in isolation.

\begin{wrapfigure}{r}{0.69\textwidth}
    \centering
    % \vspace{-24pt}
    \includegraphics[width=0.95\linewidth]{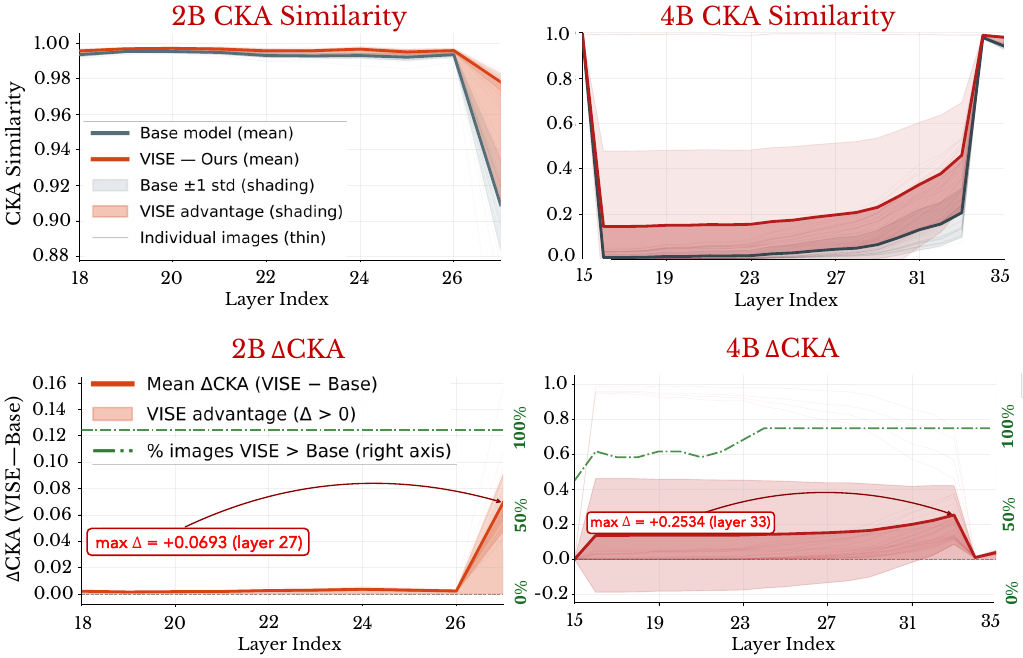}
    % \vspace{-10pt}
    \caption{\textbf{Per-layer CKA similarity between original and geometrically augmented views in Qwen3-VL decoder layers.} On 2B, VISE gains are confined to final layers (peak $\Delta=+0.069$ at layer 27) with 100\% win-rate. On 4B, gains span layers 19--33 (peak $\Delta=+0.253$), with win-rate increasing from $\sim$60\% at layer 15 to 100\% beyond layer 25.}
    \vspace{-7pt}
    \label{fig:cka}
\end{wrapfigure}

Figure~\ref{fig:cka} provides representational evidence that 
$\mathcal{R}_\text{geo}$ achieves its intended effect. We compute 
per-layer Centered Kernel Alignment (CKA) similarity~\cite{cka} between 
representations of original and geometrically augmented views for 
the base model and VISE across 100 random COCO images. On 
Qwen3-VL-2B, gains are confined to the final decoder layers, 
localizing geometric under-conditioning to the stages where 
generation decisions are formed. On Qwen3-VL-4B, the $\Delta$CKA 
advantage is distributed more broadly across the decoder, consistent 
with the scale effects in Table~\ref{tab:captioning}: concentrated 
failures yield more direct downstream gains under invariance-based 
correction.

\begin{figure}[t]
    \centering
    \includegraphics[width=\linewidth]{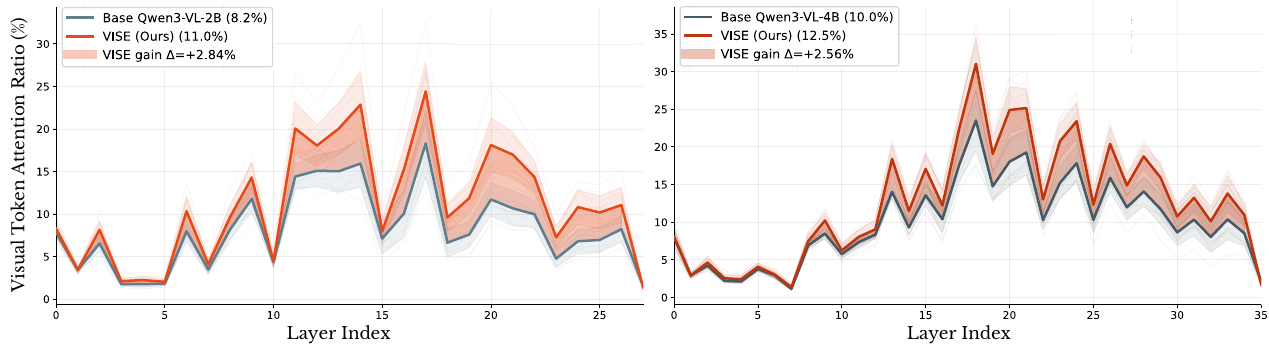}
    % \vspace{-22pt}
    \caption{\textbf{Generation-time visual attention per transformer layer for 
Base and VISE models on Qwen3-VL-2B (left) and Qwen3-VL-4B (right)}. 
VISE-trained models (orange) consistently assign more attention to 
image tokens across mid-to-late decoder layers where semantic generation 
occurs, with mean gains of $+2.84\%$ and $+2.56\%$ respectively and 
per-sample peaks of up to $+5.09\%$ in layers 15--25. The effect is 
consistent across both model scales, aligning with our claim that the semantic 
invariance reward strengthens visual conditioning during generation.}
    \label{fig:attention}
    % \vspace{-19pt}
\end{figure}

% \vspace{-17pt}

\noindent\textbf{Semantic Invariance Reward.}
Geometric consistency is a necessary but insufficient condition for
faithful visual conditioning. A model could achieve high
$\mathcal{R}_\text{geo}$ by predicting large, spatially stable
regions without its predictions being meaningfully driven by the
semantic content they enclose. Hence, we also address the complementary
dimension of visual under-conditioning: \textit{evidence sensitivity}.
A model whose responses are conditioned on the image should
recognize that removing the predicted region removes the evidence for
the queried object. If instead the decoder is shortcutting from
language priors, it would remain insensitive to that removal. We
reward the opposite: the model should judge the object as visible
when the region is intact, and as absent when it is obscured.

We do this by introducing a regional perturbation procedure termed
\textit{ghosting}. Given the predicted box $B_\text{orig}$, the
corresponding pixel region in the original image $x$ is identified
and its contents replaced by a Gaussian-blurred version with kernel
$\sigma = 25.0$, producing a ghosted image $\tilde{x}$ in which the
localized region is visually degraded while the surrounding context
is fully preserved. The model then assesses the visibility of the
queried object under $q$ on both $x$ and $\tilde{x}$ via greedy
decoding, yielding binary judgments $v = \texttt{vis}(x, q) \in
\{0, 1\}$ and $\tilde{v} = \texttt{vis}(\tilde{x}, q) \in \{0, 1\}$.
The semantic invariance reward is:
% \vspace{-7pt}
\begin{equation}
    \mathcal{R}_\text{sem} =
    \begin{cases}
        1.0 & \text{if } v = 1 \;\text{ and }\; \tilde{v} = 0 \\
        0.0 & \text{otherwise}
    \end{cases}
    \label{eq:r_sem}
\end{equation}

\noindent A prediction that is geometrically consistent but
semantically arbitrary (enclosing a region that does not contain
the queried object) receives $\mathcal{R}_\text{sem} = 0$ even if
it satisfies $\mathcal{R}_\text{geo}$, ensuring the two signals are
complementary and jointly necessary for the model to improve its
evidence binding.
Figure~\ref{fig:attention} provides direct evidence of the intended
behavioral change. We measure generation-time visual attention, which is the
fraction of attention each generated token assigns to image tokens
across transformer layers, for the base model and VISE averaged over
100 random COCO images. Unlike prefill attention, which is dominated
by the visual embedding layer, generation-time attention captures how
the decoder references visual evidence while producing each token. VISE consistently assigns more attention to visual tokens
across mid-to-late decoder layers on both model scales, indicating
that the semantic invariance reward encourages the decoder to maintain
visual conditioning throughout generation rather than reverting to
language priors after initial image encoding.

\noindent\textbf{Composite Reward and Optimization.}
The total reward combines both invariance signals as
$\mathcal{R}_t = \lambda_\text{geo} \mathcal{R}_\text{geo}
+ \lambda_\text{sem} \mathcal{R}_\text{sem}$,
where $\lambda_\text{geo} = \lambda_\text{sem} = 0.5$. At
each step, the model produces a completion $y$ containing
the predicted box coordinates for $(x, q)$. We maintain an
exponential moving average baseline $b_t \leftarrow 0.9\,b_{t-1} +
0.1\,\mathcal{R}_t$ to reduce gradient variance, giving advantage
$A_t = \mathcal{R}_t - b_t$. Letting $\Delta_t = \log p_\theta(y
\mid x, q) - \log p_\text{ref}(y \mid x, q)$ denote a KL-like
divergence proxy between the current policy and the frozen reference
$\pi_\text{ref}$, we optimize:

% \vspace{-12pt}

\begin{equation}
    \mathcal{L}(\theta) \;=\;
    -\,A_t \cdot \log p_\theta(y \mid x, q)
    \;+\; \beta_t \cdot \Delta_t
    \label{eq:loss}
\end{equation}

% \vspace{-6pt}

\noindent where the first term is a REINFORCE-style update and the
second regularizes against the reference model. We adapt the KL Coefficient $\beta_t$ dynamically to maintain a target divergence level:

% \vspace{-10pt}

\begin{equation}
    \beta_{t+1} =
    \begin{cases}
        \beta_t\,(1 + \eta) & \text{if } |\Delta_t| > \tau \\
        \beta_t\,(1 - \eta) & \text{otherwise}
    \end{cases}
    \label{eq:kl_adapt}
\end{equation}

% \vspace{-6pt}

\noindent where $\tau = 0.020$ is the target divergence budget,
$\eta = 0.10$ is the adaptation rate, and $\beta_t$ is clipped below
at $10^{-6}$. This tightens regularization when the policy drifts
beyond the target and relaxes it when updates are conservative,
providing stable self-evolution without a fixed regularization
strength.

% \vspace{-18pt}

%% file: sec/4_results.tex
\section{Experiments}

% \vspace{-10pt}

\begin{table}[t]
\input{tables/table1}
\end{table}

\noindent\textbf{Implementation Details.}
We fine-tune each base model using LoRA~\cite{lora} while keeping the vision encoder frozen. For the smaller models (2B and 4B), we use rank $r{=}16$ and $\alpha{=}32$, whereas for the larger models (8B and 32B), we increase the rank to $r{=}32$ and $\alpha$ to $64$ (dropout $0.05$ in all cases). Training uses the AdamW optimizer~\cite{adamw} with weight decay $0.01$ and gradient clipping at $1.0$. We set the learning rate to $10^{-6}$ and the KL regularization target to $0.020$ (adaptive rate $0.10$) for smaller models, and reduce them to $1.5\times10^{-7}$ and $0.004$ (adaptive rate $0.15$), respectively, for larger models. Equal reward weights of $0.5$ are applied to both the geometric and semantic invariance terms. All models are trained for 4000 steps on 8$\times$ AMD MI250X GPUs using bfloat16 precision. No question--answer pairs, annotations, metadata, or external reward models are used.

\noindent\textbf{Training Data.}
We use 4000 raw, unlabeled images sampled from the COCO
dataset~\cite{coco}, with no captions, bounding boxes, or semantic
labels retained. Spatial transformations (affine, crop, and flip) are applied online during training to generate geometric
invariance targets. Supp.~Sec.~\ref{sec:supp_validation} reports Objects365
training results showing consistent gains beyond COCO-specific image exposure.

\noindent\textbf{Baselines and Evaluation.}
We compare against five self-evolving baselines on the same base models: VisPlay~\cite{p7}, EvoLMM~\cite{p3}, and iReasoner~\cite{p1}, which are fully unsupervised and require no external rewards or annotations, and VisionZero~\cite{visionzero} (CLEVR, Chart, and Real-World [RW] variants), where only CLEVR is label-free, while Chart and Real-World use GPT-4o during dataset construction. All evaluations are run on AMD MI250X GPUs using the lmms-eval framework \cite{lmmseval}, with HuggingFace Transformers v4.38 and bfloat16 precision for consistency with training.
We evaluate on four image captioning benchmarks (COCO [2014/2017 average]~\cite{coco}, NoCaps~\cite{nocaps}, 
Flickr30k~\cite{plummer2015flickr30k}, TextCaps~\cite{sidorov2020textcaps}), twelve VQA and reasoning benchmarks (GQA~\cite{hudson2019gqa}, 
OK-VQA~\cite{marino2019ok}, VQAv2~\cite{vqav2}, AI2D~\cite{ai2d}, ChartQA~\cite{chartqa}, InfoVQA~\cite{infovqa}, ScienceQA~\cite{scienceqa}, MMMU~\cite{mmmu}, CaptionQA~\cite{yang2025captionqa}, 
RWQA~\cite{realworldqa}, ESB~\cite{du2024embspatial}, MMBench~\cite{liu2024mmbench}), and two hallucination benchmarks (POPE \cite{pope} and COCO 
Cap Chair~\cite{chair}).

% \vspace{-18pt}

\noindent\textbf{Image Captioning.}
\begin{figure}[t]
    \centering
    \includegraphics[width=0.97\linewidth]{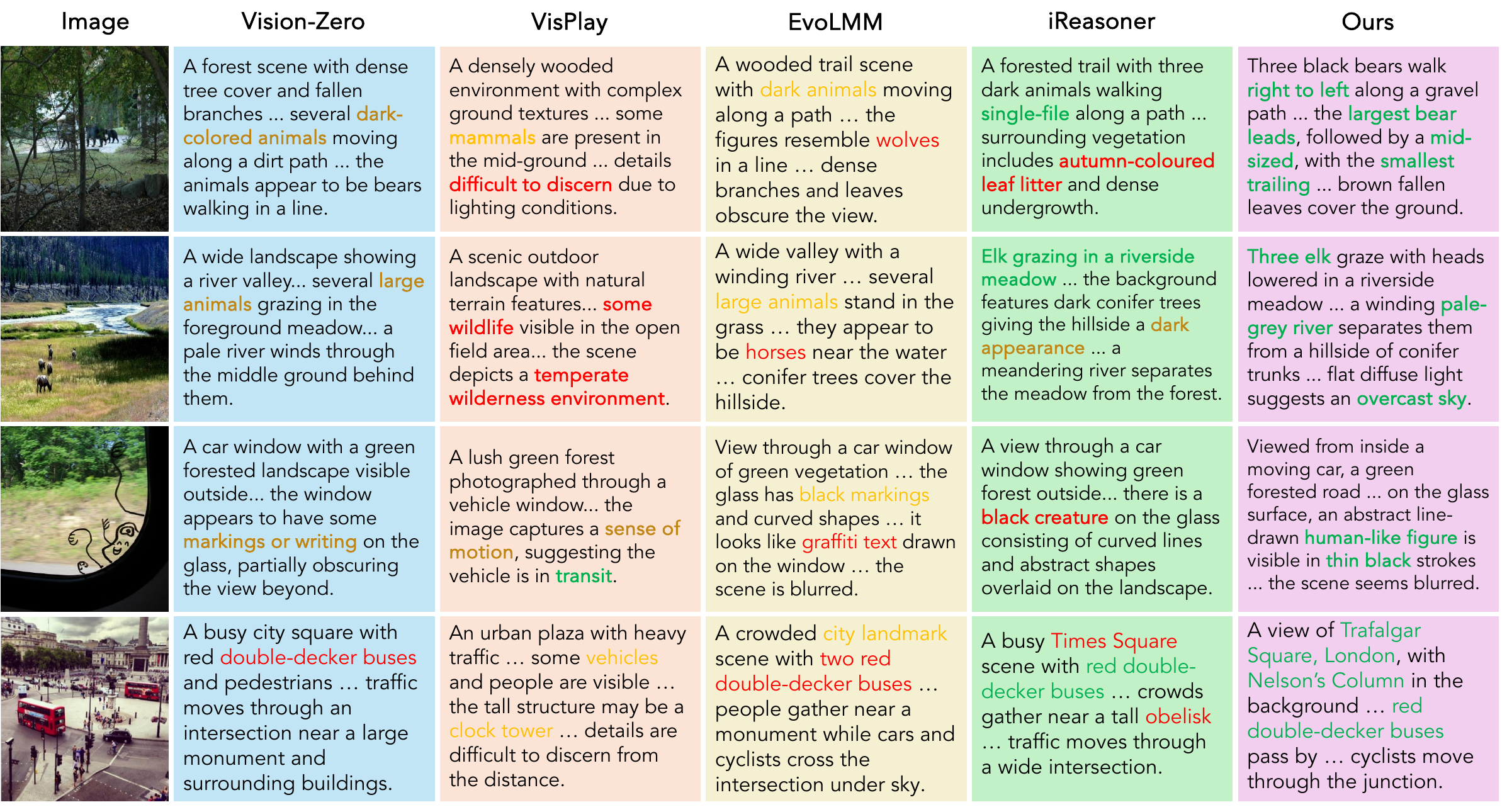}
    \vspace{-8pt}
    \caption{\textbf{Qualitative comparison of VISE and baselines on four images.}
Baselines generate either vague category-level descriptions
(“large animals,” “vehicles”) or confident hallucinations
(“wolves,” “obelisk”), reflecting reliance on language priors.
In contrast, VISE provides specific, image-grounded descriptions:
it identifies the three bears by size and position, names the elk and
river color, recognizes the human-like figure on the car window, and
correctly identifies Trafalgar Square and Nelson’s Column.
These differences are supported by CIDEr gains in
Table~\ref{tab:captioning}, indicating that invariance-based training
encourages reliance on visible evidence rather than scene-level priors.}
    \label{fig:qualitative}
    % \vspace{-10pt}
\vspace{-20pt}
\end{figure}
In Table~\ref{tab:captioning}, we compare VISE against all baselines
across four captioning benchmarks and four Qwen3-VL scales.
Consistency-based methods trained on math and scientific images show a
clear pattern at 2B: EvoLMM drops $-0.70$ CIDEr on COCO, $-0.77$ on
NoCaps, and $-0.94$ on Flickr30k, with iReasoner following a similar
trend. This persists across scales, and cannot be explained by
distribution mismatch alone; when answer agreement is the reward,
the model is never required to describe what it actually sees, so
captioning degrades. VisionZero-RW is the strongest captioning
baseline at 2B ($+4.04$ COCO CIDEr) owing to closer alignment with
natural scenes, though its Chart and CLEVR variants show smaller and
less consistent gains. VisPlay improves COCO by $+2.31$ but regresses
on NoCaps ($-0.38$) and TextCaps ($-0.09$), suggesting diversity and
difficulty rewards prioritize question complexity over visual
faithfulness. At larger scales, VisionZero variants recover further
(e.g., RW reaches $+8.41$ COCO CIDEr at 8B) while
EvoLMM and iReasoner remain inconsistent, pointing to distribution
proximity rather than visual conditioning as the reason.

VISE improves CIDEr on Qwen3-VL-2B by $+16.85$ on COCO,
$+14.73$ on NoCaps, $+16.55$ on Flickr30k, and $+19.66$ on
TextCaps with gains $4{\times}$--$7{\times}$ larger than the strongest
baseline on each benchmark and no regressions across any dataset
or scale. Gains decrease with model size ($+16.85$ at 2B to $+8.72$
at 32B on COCO), in line with larger models entering training with
stronger visual conditioning already consolidated during pretraining.
Figure~\ref{fig:qualitative} shows what these numbers look like in
practice. Where baselines either stay vague ("large animals near a
river") or commit to plausible-but-wrong details ("wolves," "obelisk"),
VISE reads the image: it catches that the animals are three bears of
different sizes walking in order, that there is a hand-drawn figure on
the car window, and that the square is Trafalgar rather than a generic
city landmark. The gap between these outputs is precisely what the
CIDEr gap in Table~\ref{tab:captioning} reflects.

\begin{table}[t]
\input{tables/table2}
\end{table}

\noindent\textbf{VQA and Reasoning.}
Table~\ref{tab:vqa} reports performance across twelve VQA and
reasoning benchmarks. The baseline results exhibit a consistent
generalization tradeoff that is not observed with VISE.
VisionZero-Chart improves ChartQA by $+0.96$ at 2B but drops
$-0.45$ on OK-VQA and $-0.50$ on CaptionQA; VisionZero-RW
shows the reverse, gaining on natural-language tasks ($+0.90$ RWQA)
but dropping on ChartQA ($-0.47$); VisionZero-CLEVR improves
structured reasoning ($+2.54$ ScienceQA, $+1.91$ InfoVQA) while
remaining inconsistent on open-ended VQA. This bidirectional pattern
suggests that domain-specific consistency training picks up the
statistical regularities of the training distribution alongside any
conditioning signal, imposing a ceiling that cannot be lifted without
changing the reward itself. EvoLMM and iReasoner follow the same
logic: strong gains on ScienceQA ($+3.59$, $+3.70$ at 2B) but drops
on OK-VQA ($-2.73$, $-2.63$) and VQAv2 ($-0.51$, $-0.43$), in line
with the captioning dips in Table~\ref{tab:captioning}.

\begin{table}[t!]
\input{tables/table3}
\end{table}

VISE improves all twelve benchmarks simultaneously at 2B, with gains
of $+4.19$ on ScienceQA, $+2.41$ on InfoVQA, $+1.75$ on MMMU, and
$+2.12$ on CaptionQA, and no regressions at any scale. At 4B, the
MMMU gain grows to $+3.72$, the largest improvement on that benchmark
across all methods and scales, indicating that stronger conditioning
particularly benefits tasks requiring multi-discipline visual
reasoning. Across all four scales, VISE exhibits no structured-versus-open-ended
tradeoff observed in the baselines. Instead of adapting to a specific
domain, the invariance reward improves the model’s visual conditioning
behavior, and these improvements generalize across task formats.

\noindent\textbf{Hallucination.}
Table~\ref{tab:hallucination} evaluates all methods on POPE and COCO
Cap Chair. VisPlay increases hallucination at 2B (Chair-I $+0.003$,
Chair-S $+0.23$), consistent with diversity rewards prioritizing
question complexity over visual fidelity. EvoLMM and iReasoner reduce
Chair scores modestly ($-0.23$, $-1.74$) but drop POPE accuracy
($-1.42$, $-1.31$) simultaneously. Sentence-level hallucinations
decline while binary object-presence reliability weakens, pointing to
inconsistent rather than substantive improvement. VisionZero-RW
is the strongest baseline (Chair-I $-2.99$, Chair-S $-4.05$),
due to broader real-world visual coverage.
VISE reduces Chair-I by $-5.00$ and Chair-S by $-5.45$ at 2B,
surpassing the best baseline by $+2.01$ and $+1.40$ while also
improving POPE by $+1.02$. This joint improvement across both metrics
is unique to VISE: penalizing confident predictions under regional
perturbation discourages generating objects that are statistically
plausible but not visually present. Gains attenuate with model size
(Chair-I $-0.37$, $-0.44$ at 8B, 32B), in-line with the
captioning trend in Table~\ref{tab:captioning}.

\noindent\textbf{Effect of Model Scale.}
Gains from VISE are largest at smaller scales and attenuate with model
size, consistent across all task groups. We attribute this to a capacity ceiling effect
in the self-evolving setting: larger models enter post-training with
stronger conditioning consolidated during pretraining and instruction
tuning, leaving less headroom for invariance-based correction. This
aligns with prior observations that self-evolving methods exhibit
diminishing returns at scale~\cite{p7}, and suggests that invariance
rewards are most effective where visual under-conditioning remains
pronounced, particularly in smaller models that have not yet developed
robust evidence-binding behavior. 

\noindent\textbf{Backbone Generalization.} Table~\ref{tab:backbone} evaluates VISE on four architecturally
diverse backbones trained under an identical setup using the same
4000 unlabeled COCO images. Captioning and hallucination gains are
consistent across families, with NoCaps showing the largest
absolute improvements. Reasoning and VQA also improve without
regressions, including on Llama-3.2-11B ($+1.31$ POPE) despite its
weaker baseline grounding, indicating that the reward is insensitive
to pretraining distribution. The consistency of improvements across
architectures confirms that the invariance reward is architecture-agnostic
and that visual under-conditioning is a general phenomenon
addressed by VISE regardless of backbone.

% \vspace{-1pt}

\begin{table}[t]
\input{tables/table4}
\end{table}

% \noindent\textbf{Backbone Generalization.}

% \vspace{-17pt}

\begin{table}[t]
\input{tables/table5}
\end{table}

\noindent\textbf{Ablation Study.}
Table~\ref{tab:ablation} isolates the contribution of each reward
component on Qwen3-VL-2B and Qwen3-VL-8B.
% \noindent\textbf{Geometric invariance only ($R_\text{geo}$).}
Training with $R_\text{geo}$ yields moderate captioning gains
($+4.83$ COCO CIDEr, $+4.33$ Flickr30k at 2B) and modest hallucination
reductions (Chair-I $-1.35$, Chair-S $-1.45$), highlighting that spatial
consistency provides a meaningful visual signal. However, because it
cannot penalize visually unsupported but plausible generations, its
improvements remain well below those of the full model.
% \noindent\textbf{Semantic invariance only ($R_\text{sem}$).}
$R_\text{sem}$ accounts for most of VISE’s gains, delivering
$+13.99$ COCO CIDEr and reducing Chair-I by $-4.15$ and Chair-S by
$-4.45$ at 2B. This is consistent with our findings that hallucination
reduction stems directly from the semantic reward design. By penalizing
confident predictions under regional perturbation, the ghosting signal
emerges as the primary driver of the captioning and hallucination gains
in Tables~\ref{tab:captioning} and~\ref{tab:hallucination}.

\noindent\textbf{Full model ($R_\text{geo} + R_\text{sem}$).}
Combining both rewards yields additional and consistent gains, with
$R_\text{geo}$ contributing complementary benefits to $R_\text{sem}$.
At 2B, the full model adds $+2.86$ COCO CIDEr and a further $-0.85$
Chair-I reduction beyond $R_\text{sem}$ alone. The same pattern holds
at 8B: while $R_\text{sem}$ dominates captioning
($+6.26$ vs.\ $+2.83$ COCO CIDEr), the combined model delivers broader
improvements across VQA and reasoning tasks. Together, these results
show that spatial consistency and evidence sensitivity address distinct
facets of visual under-conditioning and are both necessary for the full
benefits of VISE. Supp.~Sec.~\ref{sec:supp_validation} further validates this
interpretation: a random-reward control remains near base captioning
performance, confirming that the gains come from the invariance rewards rather
than fine-tuning alone.

%% file: tables/table1.tex
% \textcolor{deltaup}{Green}/\textcolor{deltadn}{red} deltas indicate
% absolute change relative to the base model.
\centering
% Table 1 — Captioning
\caption{\textbf{Evaluation results on four
image captioning benchmarks.}
C = CIDEr, M = METEOR, R = ROUGE-L. Consistency-based methods trained on math and scientific images
(EvoLMM, iReasoner) regress on captioning across all scales; EvoLMM drops $-0.70$ CIDEr on COCO and $-0.94$ on Flickr30k
at 2B, suggesting that prior-driven generation reinforces language
priors rather than correcting them. Conversely, VISE improves CIDEr
from $21.54 \rightarrow 38.39$ on COCO and $22.20 \rightarrow 41.86$
on TextCaps at 2B, with no regressions across any dataset or scale.}

% demonstrating that invariance-based rewards yield consistent
% captioning gains without caption supervision or domain-specific
% training data

% \vspace{-11pt}

\label{tab:captioning}
\resizebox{\textwidth}{!}{%
\begin{tabular}{l ccc ccc ccc ccc}
\toprule
\multirow{2}{*}{\textbf{Method}}
  & \multicolumn{3}{c}{\textbf{COCO}}
  & \multicolumn{3}{c}{\textbf{NoCaps}}
  & \multicolumn{3}{c}{\textbf{Flickr30k}}
  & \multicolumn{3}{c}{\textbf{TextCaps}} \\
\cmidrule(lr){2-4} \cmidrule(lr){5-7} \cmidrule(lr){8-10} \cmidrule(lr){11-13}
  & C & M & R  & C & M & R  & C & M & R  & C & M & R \\
\midrule

\multicolumn{13}{l}{\textit{Qwen3-VL-2B-Instruct}} \\
\hdashline
Base                 & 21.54            & 26.31            & 42.35            & 19.52            & 27.36            & 45.08            & 26.09            & 25.70            & 43.75            & 22.20            & 25.31            & 38.66            \\
VisPlay~\cite{p7}              & 23.85\up{2.31} & 
26.61\up{0.30} & 
42.65\up{0.30} & 
19.14\dn{0.38} & 
27.83\up{0.47} & 
45.43\up{0.35} & 
27.50\up{1.41} & 
25.53\dn{0.17} & 
44.20\up{0.45} & 
22.11\dn{0.09} & 
25.46\up{0.15} & 
39.13\up{0.47} \\

VisionZero (CLEVR)~\cite{visionzero} & 22.67\up{1.13} & 
26.34\up{0.03} & 
42.88\up{0.53} & 
20.16\up{0.64} & 
28.01\up{0.65} & 
44.23\dn{0.85} & 
27.78\up{1.69} & 
24.89\dn{0.81} & 
44.65\up{0.90} & 
22.82\up{0.62} & 
24.49\dn{0.82} & 
39.31\up{0.65} \\

VisionZero (Chart)~\cite{visionzero}     & 21.47\dn{0.07}& 
26.96\up{0.65} & 
43.10\up{0.75} & 
20.19\up{0.67} & 
26.74\dn{0.62}& 
44.33\dn{0.75}& 
26.98\up{0.89} & 
26.52\up{0.82} & 
42.98\dn{0.77}& 
21.41\dn{0.79}& 
25.96\up{0.65} & 
39.47\up{0.81} \\

VisionZero-RW~\cite{visionzero} & 25.58\up{4.04} & 
27.20\up{0.89} & 
43.46\up{1.11} & 
22.61\up{3.09} & 
28.24\up{0.88} & 
46.12\up{1.04} & 
29.94\up{3.85} & 
26.62\up{0.92} & 
44.69\up{0.94} & 
25.28\up{3.08} & 
\textbf{26.21}\up{0.90} & 
39.79\up{1.13} \\

EvoLMM~\cite{p3}               & 20.84\dn{0.70} & 
26.71\up{0.40} & 
41.52\dn{0.83} & 
18.75\dn{0.77} & 
26.72\dn{0.64} & 
45.86\up{0.78} & 
25.15\dn{0.94} & 
26.15\up{0.45} & 
42.91\dn{0.84} & 
23.04\up{0.84} & 
24.68\dn{0.63} & 
39.39\up{0.73} \\

iReasoner~\cite{p1}               & 20.93\dn{0.61} & 
26.75\up{0.44} & 
41.59\dn{0.76} & 
18.81\dn{0.71} & 
26.75\dn{0.61} & 
45.96\up{0.88} & 
25.23\dn{0.86} & 
26.24\up{0.54} & 
43.00\dn{0.75} & 
23.14\up{0.94} & 
24.79\dn{0.52} & 
39.47\up{0.81} \\

\rowcolor{rowblue}
\textbf{Ours}        & \textbf{38.39}\up{16.85} & 
\textbf{27.42}\up{1.11} & 
\textbf{45.86}\up{3.51} & 
\textbf{34.25}\up{14.73} & 
\textbf{28.60}\up{1.24} & 
\textbf{48.69}\up{3.61} & 
\textbf{42.64}\up{16.55} & 
\textbf{26.77}\up{1.07} & 
\textbf{47.55}\up{3.80} & 
\textbf{41.86}\up{19.66} & 
26.20\up{0.89} & 
\textbf{41.81}\up{3.15} \\

\midrule

\multicolumn{13}{l}{\textit{Qwen3-VL-4B-Instruct}} \\
\hdashline
Base                 & 27.35            & 26.20            & 42.95            & 22.36            & 28.91            & 45.95            & 31.10            & 25.64            & 44.20            & 34.54            & 25.26            & 39.70            \\
VisPlay~\cite{p7}              & 28.59\up{1.24} & 
26.41\up{0.21} & 
44.10\up{1.15} & 
25.46\up{3.10} & 
29.10\up{0.19} & 
45.72\dn{0.23} & 
34.38\up{3.28} & 
25.52\dn{0.12} & 
44.05\dn{0.15} & 
31.34\dn{3.20} & 
25.03\dn{0.23} & 
39.42\dn{0.28}\\
VisionZero-CLEVR~\cite{visionzero}     & 24.69\dn{2.66}& 
26.15\dn{0.05}& 
42.92\dn{0.03}& 
27.05\up{4.69} & 
28.71\dn{0.20}& 
45.74\dn{0.21}& 
36.29\up{5.19} & 
25.35\dn{0.29}& 
44.36\up{0.16} & 
38.08\up{3.54} & 
25.54\up{0.28} & 
39.91\up{0.21} \\
VisionZero-Chart~\cite{visionzero}     & 29.11\up{1.76} & 
26.21\up{0.01} & 
43.22\up{0.27} & 
28.12\up{5.76} & 
29.22\up{0.31} & 
46.13\up{0.18} & 
35.38\up{4.28} & 
25.81\up{0.17} & 
44.50\up{0.30} & 
32.95\dn{1.59} & 
25.47\up{0.21} & 
39.91\up{0.21} \\
VisionZero-RW~\cite{visionzero} & 31.13\up{3.78} & 
26.42\up{0.22} & 
44.70\up{1.75} & 
28.61\up{6.25} & 
29.15\up{0.24} & 
46.09\up{0.14} & 
35.67\up{4.57} & 
26.03\up{0.39} & 
44.46\up{0.26} & 
\textbf{38.89}\up{4.35} & 
25.45\up{0.19} & 
39.89\up{0.19} \\
EvoLMM~\cite{p3}                 & 30.53\up{3.18} & 
26.12\dn{0.08} & 
42.87\dn{0.08} & 
25.36\up{3.00} & 
28.61\dn{0.30} & 
45.73\dn{0.22} & 
28.16\dn{2.94} & 
25.83\up{0.19} & 
44.25\up{0.05} & 
38.29\up{3.75} & 
25.34\up{0.08} & 
39.82\up{0.12} \\
iReasoner~\cite{p1}               & 30.68\up{3.33} & 
26.25\up{0.05} & 
42.93\dn{0.02} & 
25.52\up{3.16} & 
28.73\dn{0.18} & 
45.78\dn{0.17} & 
28.24\dn{2.86} & 
25.93\up{0.29} & 
44.36\up{0.16} & 
38.41\up{3.87} & 
25.47\up{0.21} & 
39.97\up{0.27} \\
\rowcolor{rowblue}
\textbf{Ours}        & \textbf{39.65}\up{12.30} & 
\textbf{29.52}\up{3.32} & 
\textbf{46.91}\up{3.96} & 
\textbf{34.97}\up{12.61} & 
\textbf{29.60}\up{0.69} & 
\textbf{48.86}\up{2.91} & 
\textbf{37.17}\up{6.07} & 
\textbf{26.64}\up{1.00} & 
\textbf{47.82}\up{3.62} & 
38.59\up{4.05} & 
\textbf{26.43}\up{1.17} & 
\textbf{42.05}\up{2.35} \\

\midrule

\multicolumn{13}{l}{\textit{Qwen3-VL-8B-Instruct}} \\
\hdashline
Base                 & 29.01            & 27.25            & 45.72            & 24.46            & 29.91            & 47.82            & 34.02            & 26.69            & 46.55            & 36.21            & 26.28            & 41.62            \\
VisPlay~\cite{p7}              & 31.02\up{2.01} & 
27.58\up{0.33} & 
45.69\dn{0.03} & 
26.41\up{1.95} & 
29.98\up{0.07} & 
47.96\up{0.14} & 
35.01\up{0.99} & 
26.77\up{0.08} & 
46.78\up{0.23} & 
36.73\up{0.52} & 
26.24\dn{0.04} & 
41.89\up{0.27} \\
VisionZero-CLEVR~\cite{visionzero}     & 34.88\up{5.87} & 
28.36\up{1.11} & 
46.08\up{0.36} & 
29.92\up{5.46} & 
30.14\up{0.23} & 
48.44\up{0.62} & 
35.73\up{1.71} & 
26.83\up{0.14} & 
46.93\up{0.38} & 
37.21\up{1.00} & 
26.53\up{0.25} & 
41.97\up{0.35} \\
VisionZero-Chart~\cite{visionzero}     & 36.12\up{7.11} & 
28.62\up{1.37} & 
46.21\up{0.49} & 
31.47\up{7.01} & 
30.22\up{0.31} & 
48.53\up{0.71} & 
33.88\dn{0.14} & 
26.75\up{0.06} & 
46.82\up{0.27} & 
35.98\dn{0.23} & 
26.39\up{0.11} & 
41.88\up{0.26} \\
VisionZero-RW~\cite{visionzero} & 37.42\up{8.41} & 
28.71\up{1.46} & 
46.24\up{0.52} & 
33.87\up{9.41} & 
30.27\up{0.36} & 
48.59\up{0.77} & 
37.92\up{3.90} & 
27.03\up{0.34} & 
47.39\up{0.84} & 
38.03\up{1.82} & 
26.96\up{0.68} & 
42.63\up{1.01} \\
EvoLMM~\cite{p3}                 & 29.84\up{0.83} & 
27.21\dn{0.04} & 
45.78\up{0.06} & 
24.89\up{0.43} & 
29.95\up{0.04} & 
48.31\up{0.49} & 
33.74\dn{0.28} & 
26.65\dn{0.04} & 
47.02\up{0.47} & 
36.05\dn{0.16} & 
26.31\up{0.03} & 
41.73\up{0.11} \\
iReasoner~\cite{p1}            & 33.26\up{4.25} & 
28.04\up{0.79} & 
45.67\dn{0.05} & 
28.44\up{3.98} & 
30.09\up{0.18} & 
48.21\up{0.39} & 
36.34\up{2.32} & 
26.61\dn{0.08} & 
47.06\up{0.51} & 
37.48\up{1.27} & 
26.63\up{0.35} & 
42.21\up{0.59} \\
\rowcolor{rowblue}
\textbf{Ours}        & \textbf{38.49}\up{9.48} & 
\textbf{28.93}\up{1.68} & 
\textbf{46.31}\up{0.59} & 
\textbf{34.98}\up{10.52} & 
\textbf{30.31}\up{0.40} & 
\textbf{48.66}\up{0.84} & 
\textbf{38.62}\up{4.60} & 
\textbf{27.11}\up{0.42} & 
\textbf{47.61}\up{1.06} & 
\textbf{38.42}\up{2.21} & 
\textbf{27.11}\up{0.83} & 
\textbf{42.78}\up{1.16} \\

\midrule

\multicolumn{13}{l}{\textit{Qwen3-VL-32B-Instruct}} \\
\hdashline
Base                 & 33.45            & 29.72            & 49.85            & 27.74            & 33.38            & 50.02            & 38.68            & 29.73            & 49.83            & 41.25            & 29.68            & 45.85            \\
VisPlay~\cite{p7}              & 35.01\up{1.56} & 
30.01\up{0.29} & 
49.79\dn{0.06} & 
28.62\up{0.88} & 
33.51\up{0.13} & 
50.09\up{0.07} & 
39.02\up{0.34} & 
29.81\up{0.08} & 
49.96\up{0.13} & 
41.63\up{0.38} & 
29.61\dn{0.07} & 
45.96\up{0.11} \\
VisionZero-CLEVR~\cite{visionzero}     & 38.74\up{5.29} & 
30.62\up{0.90} & 
50.94\up{1.09} & 
32.41\up{4.67} & 
34.02\up{0.64} & 
50.61\up{0.59} & 
39.64\up{0.96} & 
30.31\up{0.58} & 
51.02\up{1.19} & 
42.31\up{1.06} & 
29.92\up{0.24} & 
46.21\up{0.36} \\
VisionZero-Chart~\cite{visionzero}     & 39.82\up{6.37} & 
30.81\up{1.09} & 
50.88\up{1.03} & 
33.14\up{5.40} & 
34.10\up{0.72} & 
50.66\up{0.64} & 
38.52\dn{0.16} & 
30.44\up{0.71} & 
50.74\up{0.91} & 
41.11\dn{0.14} & 
29.88\up{0.20} & 
46.12\up{0.27} \\
VisionZero-RW~\cite{visionzero} & 41.21\up{7.76} & 
30.94\up{1.22} & 
51.08\up{1.23} & 
35.12\up{7.38} & 
34.29\up{0.91} & 
50.69\up{0.67} & 
40.05\up{1.37} & 
30.76\up{1.03} & 
51.71\up{1.88} & 
42.84\up{1.59} & 
30.27\up{0.59} & 
46.41\up{0.56} \\
EvoLMM~\cite{p3}                 & 34.01\up{0.56} & 
29.63\dn{0.09} & 
49.92\up{0.07} & 
28.03\up{0.29} & 
33.44\up{0.06} & 
50.28\up{0.26} & 
38.44\dn{0.24} & 
29.69\dn{0.04} & 
50.31\up{0.48} & 
41.02\dn{0.23} & 
29.71\up{0.03} & 
45.93\up{0.08} \\
iReasoner~\cite{p1}            & 37.62\up{4.17} & 
30.48\up{0.76} & 
50.61\up{0.76} & 
31.68\up{3.94} & 
33.97\up{0.59} & 
50.53\up{0.51} & 
39.73\up{1.05} & 
30.18\up{0.45} & 
50.96\up{1.13} & 
42.19\up{0.94} & 
29.94\up{0.26} & 
46.19\up{0.34} \\
\rowcolor{rowblue}
\textbf{Ours}        & \textbf{42.17}\up{8.72} & 
\textbf{31.02}\up{1.30} & 
\textbf{51.23}\up{1.38} & 
\textbf{35.95}\up{8.21} & 
\textbf{34.41}\up{1.03} & 
\textbf{50.74}\up{0.72} & 
\textbf{40.32}\up{1.64} & 
\textbf{30.89}\up{1.16} & 
\textbf{52.02}\up{2.19} & 
\textbf{43.06}\up{1.81} & 
\textbf{30.34}\up{0.66} & 
\textbf{46.54}\up{0.69} \\

\bottomrule
\end{tabular}%
}
% \vspace{-20pt}

%% file: tables/table2.tex
\centering
% Table 2 — VQA and Reasoning
\caption{\textbf{Evaluation results on 
twelve VQA and reasoning benchmarks.} All values 
are Accuracy except CaptionQA (GPT Score). Domain-specific baselines 
exhibit a consistent generalization tradeoff: EvoLMM and iReasoner improve 
on ScienceQA ($+3.59$, $+3.70$) but drop on OK-VQA ($-2.73$, $-2.63$), 
while VisionZero variants show the reverse pattern depending on training 
domain. VISE improves all twelve benchmarks simultaneously at 2B 
($+4.19$ ScienceQA, $+2.41$ InfoVQA, $+1.75$ MMMU) with no regressions 
at any scale, demonstrating that strengthening visual grounding 
generalizes across task formats without domain-specific adaptation.}

% \vspace{-11pt}

\label{tab:vqa}
\resizebox{\textwidth}{!}{%
\begin{tabular}{l cccccccccccc}
\toprule
\textbf{Method} & \textbf{GQA} & \textbf{OK-VQA} & \textbf{VQAv2} & \textbf{AI2D} & \textbf{ChartQA} & \textbf{InfoVQA} & \textbf{ScienceQA} & \textbf{MMMU} & \textbf{CaptionQA} & \textbf{RWQA} & \textbf{ESB} & \textbf{MMBench} \\
\midrule

\multicolumn{10}{l}{\textit{Qwen3-VL-2B-Instruct}} \\
\hdashline
Base                 & 58.25            & 40.76            & 78.37            & 73.67            & 79.16            & 69.02            & 79.42            & 38.92            & 77.04        &  63.41  &  68.54 & 74.48  \\
VisPlay~\cite{p7}              & 58.65\up{0.40} &
41.15\up{0.39} &
78.59\up{0.22} &
74.14\up{0.47} &
79.46\up{0.30} &
69.96\up{0.94} &
80.74\up{1.32} &
39.27\up{0.35} &
77.35\up{0.31} &
63.62\up{0.21} &
68.56\up{0.02} & 
74.52\up{0.04} \\
VisionZero-CLEVR~\cite{visionzero}     & 58.98\up{0.73} &
\textbf{41.39}\up{0.63} &
\textbf{79.04}\up{0.67} &
75.32\up{1.65} &
79.78\up{0.62} &
70.93\up{1.91} &
81.96\up{2.54} &
39.58\up{0.66} &
77.69\up{0.65} &
63.75\up{0.34} &
69.72\up{1.18} & 
75.07\up{0.59} \\
VisionZero-Chart~\cite{visionzero}     & 58.64\up{0.39} &
40.31\dn{0.45} &
78.85\up{0.48} &
74.10\up{0.43} &
\textbf{80.12}\up{0.96} &
69.61\up{0.59} &
79.93\up{0.51} &
39.50\up{0.58} &
76.54\dn{0.50} &
63.64\up{0.23} &
68.52\dn{0.02} & 
74.51\up{0.03} \\
VisionZero-RW~\cite{visionzero} & 57.91\dn{0.34}&
40.17\dn{0.59}&
78.89\up{0.52} &
74.20\up{0.53} &
78.69\dn{0.47}&
69.58\up{0.56} &
79.90\up{0.48} &
39.27\up{0.35} &
76.69\dn{0.35} &
64.31\up{0.90} &
69.77\up{1.23} & 
74.40\dn{0.08} \\
EvoLMM~\cite{p3}                 & 59.01\up{0.76} &
38.03\dn{2.73}&
77.86\dn{0.51}&
75.78\up{2.11} &
79.80\up{0.64} &
70.69\up{1.67} &
83.01\up{3.59} &
39.08\up{0.16} &
76.73\dn{0.31} &
63.78\up{0.37} &
69.32\up{0.78} & 
74.62\up{0.14} \\
iReasoner~\cite{p1}               & 59.13\up{0.88} &
38.13\dn{2.63}&
77.94\dn{0.43}&
75.97\up{2.30} &
79.96\up{0.80} &
70.82\up{1.80} &
83.12\up{3.70} &
39.11\up{0.19} &
76.82\dn{0.22} &
63.94\up{0.53} &
69.67\up{1.13} & 
74.75\up{0.27} \\
\rowcolor{rowblue}
\textbf{Ours}        & \textbf{59.41}\up{1.16} &
41.24\up{0.48} &
78.54\up{0.17} &
\textbf{76.42}\up{2.75} &
80.08\up{0.92} &
\textbf{71.43}\up{2.41} &
\textbf{83.61}\up{4.19} &
\textbf{40.67}\up{1.75} &
\textbf{79.16}\up{2.12}&
\textbf{64.58}\up{1.17} &
\textbf{70.14}\up{1.60} & 
\textbf{76.72}\up{2.24} \\

\midrule

\multicolumn{10}{l}{\textit{Qwen3-VL-4B-Instruct}} \\
\hdashline
Base                 & 60.32            & 47.86            & 80.07            & 80.10            & 83.18            & 77.73            & 87.51            & 45.17            & 82.93      &  67.82   & 76.79  & 83.42 \\
VisPlay~\cite{p7}              & 60.06\dn{0.26}&
48.04\up{0.18} &
79.89\dn{0.18}&
80.34\up{0.24} &
82.96\dn{0.22}&
77.96\up{0.23} &
87.30\dn{0.21}&
45.42\up{0.25} &
84.16\up{1.23}&
68.67\up{0.85} &
76.11\dn{0.68} & 
83.47\up{0.05} \\
VisionZero-CLEVR~\cite{visionzero}     & 61.30\up{0.98} &
48.41\up{0.55}&
81.07\up{1.00} &
80.81\up{0.71}&
83.68\up{0.50}&
78.28\up{0.55}&
88.46\up{0.95} &
46.27\up{1.10} &
83.89\up{0.96} &
68.43\up{0.61} &
77.02\up{0.23} & 
84.83\up{1.41} \\
VisionZero-Chart~\cite{visionzero}     & 61.39\up{1.07} &
48.50\up{0.64}&
81.09\up{1.02} &
81.76\up{1.66}&
84.38\up{1.20} &
78.74\up{1.01} &
88.50\up{0.99} &
46.20\up{1.03} &
83.90\up{0.97} &
67.91\up{0.09} &
76.25\dn{0.54} & 
83.78\up{0.36} \\
VisionZero-RW~\cite{visionzero} & 60.51\up{0.19} &
48.00\up{0.14} &
80.24\up{0.17} &
80.31\up{0.21} &
82.99\dn{0.19}&
77.94\up{0.21} &
87.73\up{0.22} &
45.41\up{0.24} &
83.10\up{0.17} &
69.74\up{1.92} &
77.06\up{0.27} & 
83.32\dn{0.10} \\
EvoLMM~\cite{p3}                 & 61.20\up{0.88} &
46.72\dn{1.14} &
80.97\up{0.90} &
81.13\up{1.03} &
83.52\up{0.34}&
78.58\up{0.85} &
88.46\up{0.95} &
46.02\up{0.85} &
83.41\up{0.48} &
68.65\up{0.83} &
76.94\up{0.15} & 
84.21\up{0.79} \\
iReasoner~\cite{p1}               & 61.32\up{1.00} &
46.89\dn{0.97} &
81.10\up{1.03} &
81.22\up{1.12} &
83.72\up{0.54}&
78.76\up{1.03} &
88.62\up{1.11} &
46.13\up{0.96} &
83.52\up{0.59}&
68.92\up{1.10} &
76.98\up{0.19} & 
84.34\up{0.92} \\
\rowcolor{rowblue}
\textbf{Ours}        & \textbf{61.82}\up{1.50} &
\textbf{48.81}\up{0.95} &
\textbf{81.16}\up{1.09} &
\textbf{82.16}\up{2.06} &
\textbf{84.96}\up{1.78} &
\textbf{81.45}\up{3.72} &
\textbf{90.04}\up{2.53} &
\textbf{48.89}\up{3.72} &
\textbf{85.19}\up{2.26} &
\textbf{70.46}\up{2.64} &
\textbf{77.46}\up{0.67} &
\textbf{84.51}\up{1.09} \\

\midrule

\multicolumn{10}{l}{\textit{Qwen3-VL-8B-Instruct}} \\
\hdashline
Base                 & 61.54            & 49.84            & 81.81            & 83.31            & 84.87            & 81.23            & 90.88            & 50.12            &    85.21   &  69.28     & 77.66      &  84.71      \\
VisPlay~\cite{p7}              & 61.66\up{0.12} & 
49.93\up{0.09} & 
81.78\dn{0.03} & 
83.45\up{0.14} & 
84.79\dn{0.08} & 
81.36\up{0.13} & 
91.02\up{0.14} & 
52.61\up{2.49} & 
85.32\up{0.11} & 
69.41\up{0.13} & 
78.24\up{0.58} & 
84.62\dn{0.09} \\
VisionZero-CLEVR~\cite{visionzero}     & 62.12\up{0.58} & 
50.31\up{0.47} & 
\textbf{82.72}\up{0.91} & 
83.98\up{0.67} & 
85.18\up{0.31} & 
82.44\up{1.21} & 
92.35\up{1.47} & 
\textbf{52.89}\up{2.77} & 
86.01\up{0.80} & 
69.88\up{0.60} & 
77.96\up{0.30} & 
85.11\up{0.40} \\
VisionZero-Chart~\cite{visionzero}     & 62.05\up{0.51} & 
50.26\up{0.42} & 
82.69\up{0.88} & 
84.04\up{0.73} & 
\textbf{85.49}\up{0.62} & 
82.31\up{1.08} & 
92.41\up{1.53} & 
49.98\dn{0.14} & 
85.96\up{0.75} & 
69.95\up{0.67} & 
76.94\dn{0.72} & 
85.07\up{0.36} \\
VisionZero-RW~\cite{visionzero} & 61.71\up{0.17} & 
49.89\up{0.05} & 
81.92\up{0.11} & 
83.37\up{0.06} & 
84.94\up{0.07} & 
81.44\up{0.21} & 
91.18\up{0.30} & 
50.43\up{0.31} & 
85.38\up{0.17} & 
69.33\up{0.05} & 
78.02\up{0.36} & 
84.76\up{0.05} \\
EvoLMM~\cite{p3}                 & 61.94\up{0.40} & 
50.18\up{0.34} & 
82.38\up{0.57} & 
83.89\up{0.58} & 
85.05\up{0.18} & 
82.02\up{0.79} & 
91.96\up{1.08} & 
51.56\up{1.44} & 
85.82\up{0.61} & 
69.74\up{0.46} & 
77.51\dn{0.15} & 
85.02\up{0.31} \\
iReasoner~\cite{p1}               & 62.02\up{0.48} & 
50.24\up{0.40} & 
82.46\up{0.65} & 
83.96\up{0.65} & 
85.12\up{0.25} & 
82.15\up{0.92} & 
92.11\up{1.23} & 
51.83\up{1.71} & 
85.89\up{0.68} & 
69.81\up{0.53} & 
77.82\up{0.16} & 
85.09\up{0.38} \\
\rowcolor{rowblue}
\textbf{Ours}        & \textbf{62.43}\up{0.89} &
\textbf{50.61}\up{0.77} &
82.65\up{0.84} &
\textbf{84.10}\up{0.79} &
85.41\up{0.54} &
\textbf{82.83}\up{1.60} &
\textbf{92.81}\up{1.93} &
52.69\up{2.57} &
\textbf{86.35}\up{1.14} &
\textbf{70.03}\up{0.75} &
\textbf{78.62}\up{0.96} &
\textbf{85.46}\up{0.75} \\

\midrule

\multicolumn{10}{l}{\textit{Qwen3-VL-32B-Instruct}} \\
\hdashline

Base                 & 62.08            & 51.16            & 83.56            & 86.63            & 86.00            & 87.77            & 95.68            & 59.47            & 88.96      &  78.16   & 81.12 & 86.65   \\
VisPlay~\cite{p7}              & 62.19\up{0.11} & 
51.23\up{0.07} & 
83.52\dn{0.04} & 
86.71\up{0.08} & 
85.94\dn{0.06} & 
87.89\up{0.12} & 
95.80\up{0.12} & 
59.18\dn{0.29} & 
89.04\up{0.08} & 
78.24\up{0.08} & 
81.18\up{0.06} & 
86.58\dn{0.07} \\
VisionZero-CLEVR~\cite{visionzero}     & 62.71\up{0.63} & 
51.63\up{0.47} & 
\textbf{83.93}\up{0.37} & 
87.10\up{0.47} & 
87.02\up{1.02} & 
88.32\up{0.55} & 
96.05\up{0.37} & 
60.12\up{0.65} & 
89.74\up{0.78} & 
79.29\up{1.13} & 
81.63\up{0.51} & 
87.18\up{0.53} \\
VisionZero-Chart~\cite{visionzero}     & 62.64\up{0.56} & 
51.57\up{0.41} & 
83.90\up{0.34} & 
87.16\up{0.53} & 
86.94\up{0.94} & 
88.19\up{0.42} & 
96.11\up{0.43} & 
60.04\up{0.57} & 
89.69\up{0.73} & 
79.34\up{1.18} & 
81.55\up{0.43} & 
87.13\up{0.48} \\
VisionZero-RW~\cite{visionzero} & 62.25\up{0.17} & 
51.22\up{0.06} & 
83.61\up{0.05} & 
86.69\up{0.06} & 
86.05\up{0.05} & 
87.94\up{0.17} & 
95.84\up{0.16} & 
59.33\dn{0.14} & 
89.09\up{0.13} & 
78.31\up{0.15} & 
81.21\up{0.09} & 
86.72\up{0.07} \\
EvoLMM~\cite{p3}                 & 62.49\up{0.41} & 
51.48\up{0.32} & 
83.80\up{0.24} & 
87.02\up{0.39} & 
86.83\up{0.83} & 
88.05\up{0.28} & 
95.98\up{0.30} & 
59.92\up{0.45} & 
89.58\up{0.62} & 
78.98\up{0.82} & 
81.49\up{0.37} & 
87.02\up{0.37} \\
iReasoner~\cite{p1}               & 62.56\up{0.48} & 
51.53\up{0.37} & 
83.84\up{0.28} & 
87.08\up{0.45} & 
86.90\up{0.90} & 
88.12\up{0.35} & 
96.03\up{0.35} & 
59.85\up{0.38} & 
89.66\up{0.70} & 
79.05\up{0.89} & 
81.58\up{0.46} & 
87.08\up{0.43} \\
\rowcolor{rowblue}
\textbf{Ours}        & \textbf{62.95}\up{0.87} &
\textbf{51.76}\up{0.60} &
83.85\up{0.29} &
\textbf{87.24}\up{0.61} &
\textbf{87.27}\up{1.27} &
\textbf{88.43}\up{0.66} &
\textbf{96.21}\up{0.53} &
\textbf{62.79}\up{3.32} &
\textbf{89.96}\up{1.00} &
\textbf{79.51}\up{1.35} &
\textbf{82.39}\up{1.27} &
\textbf{87.27}\up{0.62} \\

\bottomrule
\end{tabular}%
}
% \vspace{-20pt}

%% file: tables/table3.tex
\centering
% Table 3 — Hallucination
\caption{\textbf{Evaluation results on POPE
and COCO Cap Chair hallucination benchmarks.}
$\downarrow$ indicates lower is better. EvoLMM and iReasoner modestly reduce Chair scores ($-0.23$ Chair-I at
2B) but simultaneously drop POPE accuracy ($-1.42$, $-1.31$),
indicating inconsistent visual grounding. In contrast, VISE reduces
Chair-I from $13.21 \rightarrow 8.21$ ($-5.00$) and Chair-S from
$45.96 \rightarrow 40.51$ ($-5.45$) at 2B, surpassing the strongest
baseline by $+2.01$ and $+1.40$, while also improving POPE by
$+1.02$.}

% \vspace{-11pt}

\label{tab:hallucination}
\resizebox{\textwidth}{!}{%
\begin{tabular}{l cccc ccc}
\toprule
\multirow{2}{*}{\textbf{Method}}
  & \multicolumn{4}{c}{\textbf{POPE}}
  & \multicolumn{3}{c}{\textbf{COCO Cap Chair}} \\
\cmidrule(lr){2-5} \cmidrule(lr){6-8}
  & Acc & F1 & Prec & Recall & Chair-I$\downarrow$ & Chair-S$\downarrow$ & Cap Recall \\
\midrule

\multicolumn{8}{l}{\textit{Qwen3-VL-2B-Instruct}} \\
\hdashline
Base                 & 89.01            & 88.37            & 92.45            & 84.63            & 13.2133           & 45.9601           & 72.0939           \\
VisPlay~\cite{p7}              & 89.32\up{0.31} & 
88.72\up{0.35} & 
92.86\up{0.41} & 
84.93\up{0.30} & 
13.2168\chairup{0.0035}& 
46.1869\chairup{0.2268} & 
71.3239\dn{0.7700} \\
VisionZero-CLEVR~\cite{visionzero}     & 89.65\up{0.64} & 
89.01\up{0.64} & 
93.09\up{0.64} & 
85.27\up{0.64} & 
11.9074\chairdn{1.3059}& 
44.9661\chairdn{0.9940}& 
72.3015\up{0.2076} \\
VisionZero-Chart~\cite{visionzero}     & 89.35\up{0.34} & 
88.75\up{0.38} & 
91.86\dn{0.59}& 
\textbf{85.84}\up{1.21}& 
12.4823\chairdn{0.7310}& 
43.3741\chairdn{2.5860}& 
71.8795\dn{0.2144} \\
VisionZero-RW~\cite{visionzero} & 88.70\dn{0.31}& 
87.84\dn{0.53}& 
93.05\up{0.60} & 
83.18\dn{1.45} & 
10.2192\chairdn{2.9941}& 
41.9136\chairdn{4.0465}& 
72.0988\up{0.0049} \\
EvoLMM~\cite{p3}                 & 87.59\dn{1.42}& 
88.51\up{0.14} & 
92.05\dn{0.40}& 
85.17\up{0.54} & 
12.9851\chairdn{0.2282} & 
44.2238\chairdn{1.7363}& 
70.9924\dn{1.1015} \\
iReasoner~\cite{p1}                 & 87.70\dn{1.31}& 
88.56\up{0.19} & 
92.12\dn{0.33}& 
85.26\up{0.63} & 
12.9835\chairdn{0.2298} & 
44.2205\chairdn{1.7396}& 
70.9931\dn{1.1008} \\
\rowcolor{rowblue}
\textbf{Ours}        & \textbf{90.03}\up{1.02} & 
\textbf{89.22}\up{0.85} & 
\textbf{93.13}\up{0.68} & 
85.63\up{1.00} & 
\textbf{8.2132}\chairdn{5.0001} & 
\textbf{40.5097}\chairdn{5.4504} & 
\textbf{72.3145}\up{0.2206} \\
\midrule

\multicolumn{8}{l}{\textit{Qwen3-VL-4B-Instruct}} \\
\hdashline
Base                 & 89.73            & 88.45            & 92.57            & 84.68            & 12.9141           & 44.5792           & 74.8582           \\
VisPlay~\cite{p7}              & 89.42\dn{0.31} &
88.92\up{0.47} &
92.89\up{0.32} &
84.99\up{0.31} &
13.2158\chairup{0.3017}& 
46.1849\chairup{1.6057} & 
74.3239\dn{0.5343}  \\
VisionZero-CLEVR~\cite{visionzero}     & 89.82\up{0.09} & 
88.02\dn{0.43} & 
93.71\up{1.14} & 
82.98\dn{1.70} & 
12.6038\chairdn{0.3103} & 
45.3978\chairup{0.8186} & 
74.8559\dn{0.0023} \\
VisionZero-Chart~\cite{visionzero}     & 88.75\dn{0.98} & 
88.77\up{0.32} & 
92.79\up{0.22} & 
84.93\up{0.25}& 
12.2069\chairdn{0.7072}& 
44.4025\chairdn{0.1767}& 
75.1604\up{0.3022} \\
VisionZero-RW~\cite{visionzero} & 89.67\dn{0.06} & 
\textbf{89.25}\up{0.80} & 
\textbf{93.80}\up{1.23} & 
85.12\up{0.44} & 
11.9079\chairdn{1.0062}& 
43.4024\chairdn{1.1768}& 
74.8604\up{0.0022} \\
EvoLMM~\cite{p3}                 & 89.53\dn{0.20}& 
88.56\up{0.11} & 
93.07\up{0.50} & 
84.39\dn{0.29} & 
12.2031\chairdn{0.7110} & 
44.9048\chairup{0.3256}& 
75.0184\up{0.1602} \\
iReasoner~\cite{p1}               & 89.64\dn{0.09}& 
88.59\up{0.14} & 
93.15\up{0.58} & 
84.46\dn{0.22} & 
12.2024\chairdn{0.7117} & 
44.9035\chairup{0.3243}& 
75.0196\up{0.1614} \\
\rowcolor{rowblue}
\textbf{Ours}        & \textbf{89.86}\up{0.13} & 
88.98\up{0.53}& 
92.47\dn{0.10}& 
\textbf{85.74}\up{1.06} & 
\textbf{11.9041}\chairdn{1.0100}& 
\textbf{43.0197}\chairdn{1.5595}& 
\textbf{75.7404}\up{0.8822} \\

\midrule

\multicolumn{8}{l}{\textit{Qwen3-VL-8B-Instruct}} \\
\hdashline
Base                 & 89.91            & 88.61            & 92.71            & 84.86            & 11.2047           & 43.4237           & 75.4236          \\
VisPlay~\cite{p7}               & 90.23\up{0.32} & 
89.15\up{0.54} & 
\textbf{92.73}\up{0.02} & 
85.84\up{0.98} & 
10.9364\chairdn{0.2683} & 
42.0186\chairdn{1.4051} & 
75.7562\up{0.3326} \\
VisionZero-CLEVR~\cite{visionzero}              & 90.18\up{0.27} & 
89.06\up{0.45} & 
92.72\up{0.01} & 
85.68\up{0.81} & 
10.9829\chairdn{0.2218} & 
42.2564\chairdn{1.1673} & 
75.7014\up{0.2778} \\
VisionZero-Chart~\cite{visionzero}             & 89.95\up{0.04} & 
88.69\up{0.08} & 
92.70\dn{0.01} & 
85.01\up{0.15} & 
11.1628\chairdn{0.0419} & 
43.2015\chairdn{0.2222} & 
75.4728\up{0.0492} \\
VisionZero-RW~\cite{visionzero}     & 90.02\up{0.11} & 
88.80\up{0.19} & 
92.72\up{0.01} & 
85.20\up{0.34} & 
11.1094\chairdn{0.0953} & 
42.9624\chairdn{0.4613} & 
75.5316\up{0.1080} \\
EvoLMM~\cite{p3}   & 90.13\up{0.22} & 
88.98\up{0.37} & 
\textbf{92.73}\up{0.02} & 
85.52\up{0.66} & 
11.0187\chairdn{0.1860} & 
42.4872\chairdn{0.9365} & 
75.6441\up{0.2205} \\
iReasoner~\cite{p1}     & 90.07\up{0.16} & 
88.89\up{0.28} & 
92.72\up{0.01} & 
85.36\up{0.50} & 
11.0712\chairdn{0.1335} & 
42.7358\chairdn{0.6879} & 
75.5893\up{0.1657} \\

\rowcolor{rowblue}
\textbf{Ours}        & \textbf{90.32}\up{0.41} & 
\textbf{89.32}\up{0.71} & 
\textbf{92.73}\up{0.02} & 
\textbf{86.15}\up{1.29} & 
\textbf{10.8375}\chairdn{0.3672} & 
\textbf{41.5336}\chairdn{1.8901} & 
\textbf{75.8369}\up{0.4133} \\

\midrule

\multicolumn{8}{l}{\textit{Qwen3-VL-32B-Instruct}} \\
\hdashline
Base                 & 90.35            & 89.45            & 92.96            & 86.20            & 10.8543         & 42.6321           & 76.5346           \\
VisPlay~\cite{p7}    & 90.45\up{0.10} & 
89.57\up{0.12} & 
93.07\up{0.11} & 
86.32\up{0.12} & 
10.7564\chairdn{0.0979} & 
42.3891\chairdn{0.2430} & 
76.6943\up{0.1597} \\
VisionZero-CLEVR~\cite{visionzero}               & 90.73\up{0.38} & 
89.86\up{0.41} & 
93.39\up{0.43} & 
86.59\up{0.39} & 
10.5468\chairdn{0.3075} & 
41.9264\chairdn{0.7057} & 
77.2193\up{0.6847} \\
VisionZero-Chart~\cite{visionzero}     & 90.51\up{0.16} & 
89.64\up{0.19} & 
93.14\up{0.18} & 
86.39\up{0.19} & 
10.7085\chairdn{0.1458} & 
42.2678\chairdn{0.3643} & 
76.8125\up{0.2779} \\
VisionZero-RW~\cite{visionzero}               & 90.66\up{0.31} & 
89.80\up{0.35} & 
93.31\up{0.35} & 
86.54\up{0.34} & 
10.5994\chairdn{0.2549} & 
42.0032\chairdn{0.6289} & 
77.0816\up{0.5470} \\
EvoLMM~\cite{p3}   & 90.59\up{0.24} & 
89.72\up{0.27} & 
93.23\up{0.27} & 
86.46\up{0.26} & 
10.6521\chairdn{0.2022} & 
42.1346\chairdn{0.4975} & 
76.9348\up{0.4002} \\
iReasoner~\cite{p1}              & 90.39\up{0.04} & 
89.49\up{0.04} & 
92.95\dn{0.01} & 
86.28\up{0.08} & 
10.8129\chairdn{0.0414} & 
42.5184\chairdn{0.1137} & 
76.5812\up{0.0466} \\

\rowcolor{rowblue}
\textbf{Ours}        & \textbf{90.81}\up{0.46} & 
\textbf{89.92}\up{0.47} & 
\textbf{93.45}\up{0.49} & 
\textbf{86.65}\up{0.45} & 
\textbf{10.4173}\chairdn{0.4370} & 
\textbf{41.8735}\chairdn{0.7586} & 
\textbf{77.4621}\up{0.9275} \\
\bottomrule
\end{tabular}%
}
% \vspace{-20pt}

%% file: tables/table4.tex
\centering
% Table 4 — Model Agnosticity
\caption{\textbf{Effectiveness of VISE across four architecturally diverse LMM
backbones.}
Captioning gains are consistent across all families, with COCO CIDEr
improving by $+9.48$, $+9.01$, $+7.65$, and $+6.44$, and NoCaps
showing the largest absolute gains ($+10.52$, $+10.16$, $+8.67$,
$+6.25$). Hallucination and VQA improvements are similarly uniform,
with no regressions across architectures., showing that
the invariance reward is architecture-agnostic and that visual
prior-driven generation is a general phenomenon addressed by VISE
regardless of backbone.}
% \vspace{-11pt}
\label{tab:backbone}
\setlength{\tabcolsep}{4pt}
\resizebox{\textwidth}{!}{%
\begin{tabular}{l cccc ccc cc cccc}
\toprule
\multirow{2}{*}{\textbf{Model}}
  & \multicolumn{4}{c}{\textbf{Captioning}}
  & \multicolumn{3}{c}{\textbf{Hallucination}}
  & \multicolumn{2}{c}{\textbf{VQA}}
  & \multicolumn{4}{c}{\textbf{Reasoning}} \\
\cmidrule(lr){2-5} \cmidrule(lr){6-8} \cmidrule(lr){9-10} \cmidrule(lr){11-14}
  & COCO & NoCaps & Flickr30k & TextCaps
  & Chair-I$\downarrow$ & Chair-S$\downarrow$ & POPE\textsubscript{F1}
  & GQA & CaptionQA
  & AI2D & ChartQA & InfoVQA & ScienceQA \\
\midrule
\multicolumn{14}{l}{\textit{Qwen3-VL-8B-Instruct}} \\
\hdashline
Base
  & $29.01$ & $24.46$ & $34.02$ & $36.21$
  & $11.20$ & $43.42$ & $88.61$
  & $61.54$ & $85.21$
  & $83.31$ & $84.87$ & $81.23$ & $90.88$ \\
\rowcolor{cyan!5}
\textbf{Ours}
  & $\textbf{38.49}$\up{9.48} & $\textbf{34.98}$\up{10.52} & $\textbf{38.62}$\up{4.60} & $\textbf{38.42}$\up{2.21}
  & $\textbf{10.84}$\chairdn{0.36} & $\textbf{41.53}$\chairdn{1.89} & $\textbf{89.32}$\up{0.71}
  & $\textbf{62.43}$\up{0.89} & $\textbf{86.35}$\up{1.14}
  & $\textbf{84.10}$\up{0.79} & $\textbf{85.41}$\up{0.54} & $\textbf{82.83}$\up{1.60} & $\textbf{92.81}$\up{1.93} \\
\midrule
\multicolumn{14}{l}{\textit{InternVL3-8B-Instruct}} \\
\hdashline
Base
  & $28.43$ & $23.81$ & $33.47$ & $35.62$
  & $11.43$ & $43.87$ & $90.83$
  & $61.12$ & $84.73$
  & $83.19$ & $82.40$ & $68.77$ & $97.19$ \\
\rowcolor{cyan!5}
\textbf{Ours}
  & $\textbf{37.44}$\up{9.01} & $\textbf{33.97}$\up{10.16} & $\textbf{37.53}$\up{4.06} & $\textbf{37.94}$\up{2.32}
  & $\textbf{11.14}$\chairdn{0.29} & $\textbf{42.31}$\chairdn{1.56} & $\textbf{91.64}$\up{0.81}
  & $\textbf{61.94}$\up{0.82} & $\textbf{85.96}$\up{1.23}
  & $\textbf{83.61}$\up{0.42} & $\textbf{82.76}$\up{0.36} & $\textbf{69.31}$\up{0.54} & $\textbf{97.77}$\up{0.58} \\
\midrule
\multicolumn{14}{l}{\textit{Gemma3-12B-It}} \\
\hdashline
Base
  & $22.18$ & $18.94$ & $27.53$ & $24.06$
  & $13.57$ & $46.83$ & $80.65$
  & $56.38$ & $74.92$
  & $79.05$ & $55.64$ & $50.69$ & $83.41$ \\
\rowcolor{cyan!5}
\textbf{Ours}
  & $\textbf{29.83}$\up{7.65} & $\textbf{27.61}$\up{8.67} & $\textbf{31.47}$\up{3.94} & $\textbf{25.93}$\up{1.87}
  & $\textbf{13.28}$\chairdn{0.29} & $\textbf{45.37}$\chairdn{1.46} & $\textbf{81.69}$\up{1.04}
  & $\textbf{57.14}$\up{0.76} & $\textbf{75.88}$\up{0.96}
  & $\textbf{79.38}$\up{0.33} & $\textbf{55.97}$\up{0.33} & $\textbf{50.94}$\up{0.25} & $\textbf{83.89}$\up{0.48} \\
\midrule
\multicolumn{14}{l}{\textit{Llama-3.2-11B-Vision-Instruct}} \\
\hdashline
Base
  & $16.37$ & $14.22$ & $21.84$ & $18.53$
  & $16.24$ & $52.17$ & $75.83$
  & $51.63$ & $67.44$
  & $46.44$ & $29.24$ & $56.69$ & $56.87$ \\
\rowcolor{cyan!5}
\textbf{Ours}
  & $\textbf{22.81}$\up{6.44} & $\textbf{20.47}$\up{6.25} & $\textbf{25.63}$\up{3.79} & $\textbf{20.11}$\up{1.58}
  & $\textbf{15.97}$\chairdn{0.27} & $\textbf{51.03}$\chairdn{1.14} & $\textbf{77.14}$\up{1.31}
  & $\textbf{52.29}$\up{0.66} & $\textbf{68.31}$\up{0.87}
  & $\textbf{46.71}$\up{0.27} & $\textbf{29.48}$\up{0.24} & $\textbf{56.93}$\up{0.24} & $\textbf{57.43}$\up{0.56} \\
\bottomrule
\end{tabular}
}
% \vspace{-20pt}

%% file: tables/table5.tex
\centering
% Table 5 — Ablation
\caption{\textbf{Ablation study of VISE on the contribution of each invariance reward component.}
R\textsubscript{geo}: geometric invariance reward only.
R\textsubscript{sem}: semantic invariance reward only.
COCO CIDEr is averaged over the 2014/2017 val splits.
R\textsubscript{geo} yields moderate captioning gains ($+4.83$ COCO CIDEr at 2B) and modest hallucination reductions (Chair-I $-1.35$), indicating that spatial consistency provides a grounding signal but does not penalize evidence-agnostic generation on its own.
R\textsubscript{sem} accounts for most of the improvement ($+13.99$ COCO CIDEr, Chair-I $-4.15$ at 2B), showing that the ghosting-based signal drives the majority of captioning and hallucination gains.
Combining both produces consistent additional improvements, with the full model adding $+2.86$ CIDEr and $-0.85$ Chair-I beyond R\textsubscript{sem}, showing complementary benefits from spatial consistency and evidence sensitivity.}
% \vspace{-11pt}
\label{tab:ablation}
\resizebox{\textwidth}{!}{%
\begin{tabular}{l cccc ccc cc cccc}
\toprule
\multirow{2}{*}{\textbf{Method}}
  & \multicolumn{4}{c}{\textbf{Captioning}}
  & \multicolumn{3}{c}{\textbf{Hallucination}}
  & \multicolumn{2}{c}{\textbf{VQA}}
  & \multicolumn{4}{c}{\textbf{Reasoning}} \\
\cmidrule(lr){2-5} \cmidrule(lr){6-8} \cmidrule(lr){9-10} \cmidrule(lr){11-14}
  & COCO & NoCaps & Flickr30k & TextCaps
  & Chair-I$\downarrow$ & Chair-S$\downarrow$ & POPE
  & GQA & CaptionQA
  & AI2D & ChartQA & InfoVQA & ScienceQA \\
\midrule
\multicolumn{14}{l}{\textit{Qwen3-VL-2B-Instruct}} \\
\hdashline
Base
  & 21.54            & 19.52            & 26.09            & 22.20
  & 13.21            & 45.96            & 89.01
  & 58.25            & 77.04
  & 73.67            & 79.16            & 69.02            & 79.42            \\
R\textsubscript{geo} only
  &  26.37\up{4.83}  & 23.07\up{3.55}  & 30.42\up{4.33}  & 27.42\up{5.22}
  & 11.86\chairdn{1.35}  & 44.51\chairdn{1.45}  & 89.29\up{0.28}
  & 58.56\up{0.31}  & 77.61\up{0.57}
  & 74.21\up{0.54}  & 79.53\up{0.37}  & 69.71\up{0.69}  & 80.18\up{0.76}  \\
R\textsubscript{sem} only
  &  35.53\up{13.99}  & 31.75\up{12.23}  & 39.83\up{13.74}  & 38.52\up{16.32}
  & 9.06\chairdn{4.15}   & 41.51\chairdn{4.45}  & 89.86\up{0.85}
  & 59.21\up{0.96}  & 78.80\up{1.76}
  & 75.08\up{1.41}  & 79.82\up{0.66}  & 70.61\up{1.59}  & 81.94\up{2.52}  \\
\rowcolor{rowblue}
\textbf{Full (Ours)}
  & 38.39\up{16.85} & 34.25\up{14.73} & 42.64\up{16.55} & 41.86\up{19.66}
  & 8.21\chairdn{5.00} & 40.51\chairdn{5.45} & 90.03\up{1.02}
  & 59.41\up{1.16} & 79.16\up{2.12}
  & 76.42\up{2.75} & 80.08\up{0.92} & 71.43\up{2.41} & 83.61\up{4.19}  \\
\midrule
\multicolumn{14}{l}{\textit{Qwen3-VL-8B-Instruct}} \\
\hdashline
Base
  & 29.01            & 24.46            & 34.02            & 36.21
  & 11.20            & 43.42            & 89.91
  & 61.54            & 85.21
  & 83.31            & 84.87            & 81.23            & 90.88            \\
R\textsubscript{geo} only
  & 31.84\up{2.83}   & 27.12\up{2.66}   & 36.18\up{2.16}   & 37.93\up{1.72}
  & 11.02\chairdn{0.18}  & 42.88\chairdn{0.54}  & 89.99\up{0.08}
  & 61.79\up{0.25}   & 85.52\up{0.31}
  & 83.82\up{0.51}   & 85.21\up{0.34}   & 83.02\up{1.79}  & 92.83\up{1.95}  \\
R\textsubscript{sem} only
  & 35.27\up{6.26}   & 31.43\up{6.97}   & 37.41\up{3.39}   & 37.18\up{0.97}
  & 10.91\chairdn{0.29}  & 42.16\chairdn{1.26}  & 90.11\up{0.20}
  & 62.08\up{0.54}   & 85.94\up{0.73}
  & 83.96\up{0.65}   & 85.33\up{0.46}   & 83.11\up{1.88}  & 92.85\up{1.97}  \\
\rowcolor{rowblue}
\textbf{Full (Ours)}
  & 38.49\up{9.48}   & 34.98\up{10.52}  & 38.62\up{4.60}   & 38.42\up{2.21}
  & 10.84\chairdn{0.36} & 41.53\chairdn{1.89} & 90.32\up{0.41}
  & 62.43\up{0.89}   & 86.35\up{1.14}
  & 84.10\up{0.79}   & 85.41\up{0.54}   & 82.83\up{1.60}  & 92.81\up{1.93}  \\
\bottomrule
\end{tabular}%
}
% \vspace{-20pt}

%% file: sec/5_conclusion.tex
\section{Conclusion}
We introduced VISE, a fully unsupervised self-evolving framework for 
large multimodal models that directly addresses visual under-conditioning 
without relying on human annotations, external reward models, or 
multi-role formulations. By training within a single model using 
geometric consistency under spatial transformations and semantic 
sensitivity under regional perturbation, VISE encourages the decoder to 
pay more attention to visual tokens rather than relying on statistical 
language priors. Our experiments show consistent gains across captioning, 
VQA, reasoning, and hallucination benchmarks with no task tradeoffs, and 
these improvements hold across four model scales and four architecturally 
diverse backbones. Ablations confirm that semantic invariance drives most 
gains while geometric invariance contributes complementary improvements, 
together covering distinct dimensions of the failure mode. These results 
highlight that answer-consistency rewards are insufficient for genuine 
visual improvement: directly increasing attention to visual tokens during 
decoding is both necessary and sufficient to produce broad, robust gains. 
This work suggests a promising direction for self-evolving multimodal 
training, shifting the objective from output agreement to 
evidence-conditioned generation.

\section{Acknowledgement}
The computations were enabled by resources provided by  LUMI hosted by CSC (Finland) and LUMI consortium, and by Berzelius resource provided by the Knut and Alice Wallenberg Foundation at the NSC.

%% file: supplementary.tex
\begin{center}
    {\Large\textbf{Supplementary Material}}
\end{center}
\vspace{1em}

% S-numbering for sections, figures, and tables
\renewcommand{\thesection}{S\arabic{section}}
\renewcommand{\thefigure}{S\arabic{figure}}
\renewcommand{\thetable}{S\arabic{table}}
\setcounter{section}{0}
\setcounter{figure}{0}
\setcounter{table}{0}

\section{Hyperparameter Sensitivity}

Table~\ref{tab:sensitivity} reports how VISE performs under different
reward weight ratios and KL divergence targets on Qwen3-VL-2B and 8B.
The results are stable across both axes. Varying the
$\lambda$ ratio between the geometric and semantic rewards
($0.75/0.25$, $0.50/0.50$, $0.25/0.75$) produces differences well
under 0.5 CIDEr on captioning and under 0.3 on Chair-I, with no
clear winner across all benchmarks simultaneously. The equal-weight default lies near the center of this range and was selected on principled grounds rather than tuned on evaluation data. One
pattern worth noting is that shifting weight toward $\mathcal{R}_\text{sem}$
tends to slightly improve reasoning metrics while marginally worsening
hallucination, and the reverse holds for $\mathcal{R}_\text{geo}$,
which is consistent with the complementary roles each reward plays
as described in the ablation.

The KL target $\tau$ shows similarly low sensitivity. Tightening to
$\tau = 0.010$ slightly improves hallucination metrics (Chair-I
$-0.29$ relative to default at 2B) at the cost of marginally reduced
captioning gains, while relaxing to $\tau = 0.050$ has the opposite
effect, allowing slightly more policy drift and producing small
improvements on reasoning at the expense of hallucination. Neither
direction degrades performance meaningfully, suggesting the adaptive
KL mechanism is doing its job of keeping updates stable regardless
of the target value. Taken together, these results indicate that
VISE is not sensitive to precise hyperparameter choices within
reasonable ranges, and that the reported gains are a robust
property of the invariance-based training objective rather than
an artifact of careful tuning.

\begin{table*}[h]
\centering
\caption{Hyperparameter sensitivity analysis on Qwen3-VL-2B and Qwen3-VL-8B.
The default configuration ($\lambda_\text{geo}{=}\lambda_\text{sem}{=}0.5$,
$\tau{=}0.020$) is highlighted. All other training settings are identical
to the main experiments. $\downarrow$ indicates lower is better.}
\setlength{\tabcolsep}{4pt}
\resizebox{\textwidth}{!}{%
\begin{tabular}{ll cccc ccc cc cccc}
\toprule
\multirow{2}{*}{\textbf{Variant}} & \multirow{2}{*}{\textbf{Config}}
  & \multicolumn{4}{c}{\textbf{Captioning}}
  & \multicolumn{3}{c}{\textbf{Hallucination}}
  & \multicolumn{2}{c}{\textbf{VQA}}
  & \multicolumn{4}{c}{\textbf{Reasoning}} \\
\cmidrule(lr){3-6} \cmidrule(lr){7-9} \cmidrule(lr){10-11} \cmidrule(lr){12-15}
  & & COCO & NoCaps & Flickr30k & TextCaps
  & Chair-I$\downarrow$ & Chair-S$\downarrow$ & POPE\textsubscript{acc}
  & GQA & CaptionQA
  & AI2D & ChartQA & InfoVQA & ScienceQA \\
\midrule
\multicolumn{15}{l}{\textit{Qwen3-VL-2B-Instruct}} \\
\hdashline
\multirow{3}{*}{$\lambda$ ratio}
  & $\lambda_\text{geo}{=}0.75,\;\lambda_\text{sem}{=}0.25$
  & 38.18 & 34.06 & 42.42 & 41.72
  & 8.05 & 40.12 & 90.10
  & 59.25 & 78.95
  & 76.08 & 79.86 & 71.18 & 83.24 \\
\rowcolor{cyan!5}
  & $\lambda_\text{geo}{=}0.50,\;\lambda_\text{sem}{=}0.50$ \textbf{(default)}
  & 38.39 & 34.25 & 42.64 & 41.86
  & 8.21 & 40.51 & 90.03
  & 59.41 & 79.16
  & 76.42 & 80.08 & 71.43 & 83.61 \\
  & $\lambda_\text{geo}{=}0.25,\;\lambda_\text{sem}{=}0.75$
  & 38.33 & 34.20 & 42.57 & 41.83
  & 8.40 & 40.83 & 89.95
  & 59.34 & 79.10
  & 76.60 & 80.30 & 71.38 & 83.80 \\
\midrule
\multirow{3}{*}{KL target $\tau$}
  & $\tau = 0.010$
  & 38.31 & 34.16 & 42.56 & 41.80
  & 7.92 & 40.04 & 90.08
  & 59.30 & 79.04
  & 76.20 & 79.92 & 71.25 & 83.35 \\
\rowcolor{cyan!5}
  & $\tau = 0.020$ \textbf{(default)}
  & 38.39 & 34.25 & 42.64 & 41.86
  & 8.21 & 40.51 & 90.03
  & 59.41 & 79.16
  & 76.42 & 80.08 & 71.43 & 83.61 \\
  & $\tau = 0.050$
  & 38.34 & 34.21 & 42.60 & 41.84
  & 8.63 & 41.12 & 89.86
  & 59.38 & 79.12
  & 76.57 & 80.22 & 71.62 & 83.78 \\
\midrule
\multicolumn{15}{l}{\textit{Qwen3-VL-8B-Instruct}} \\
\hdashline
\multirow{3}{*}{$\lambda$ ratio}
  & $\lambda_\text{geo}{=}0.75,\;\lambda_\text{sem}{=}0.25$
  & 38.36 & 34.86 & 38.51 & 38.30
  & 10.56 & 41.05 & 90.38
  & 62.29 & 86.20
  & 83.92 & 85.22 & 82.66 & 92.55 \\
\rowcolor{cyan!5}
  & $\lambda_\text{geo}{=}0.50,\;\lambda_\text{sem}{=}0.50$ \textbf{(default)}
  & 38.49 & 34.98 & 38.62 & 38.42
  & 10.84 & 41.53 & 90.32
  & 62.43 & 86.35
  & 84.10 & 85.41 & 82.83 & 92.81 \\
  & $\lambda_\text{geo}{=}0.25,\;\lambda_\text{sem}{=}0.75$
  & 38.44 & 34.95 & 38.58 & 38.39
  & 11.10 & 41.86 & 90.20
  & 62.41 & 86.33
  & 84.26 & 85.55 & 82.79 & 93.02 \\
\midrule
\multirow{3}{*}{KL target $\tau$}
  & $\tau = 0.010$
  & 38.45 & 34.96 & 38.60 & 38.40
  & 10.62 & 41.12 & 90.36
  & 62.38 & 86.30
  & 84.02 & 85.33 & 82.74 & 92.70 \\
\rowcolor{cyan!5}
  & $\tau = 0.020$ \textbf{(default)}
  & 38.49 & 34.98 & 38.62 & 38.42
  & 10.84 & 41.53 & 90.32
  & 62.43 & 86.35
  & 84.10 & 85.41 & 82.83 & 92.81 \\
  & $\tau = 0.050$
  & 38.46 & 34.97 & 38.61 & 38.41
  & 11.12 & 41.98 & 90.18
  & 62.41 & 86.33
  & 84.24 & 85.53 & 83.01 & 93.05 \\
\bottomrule
\end{tabular}%
}
\label{tab:sensitivity}
\end{table*}

\section{Additional Validation Experiments}
\label{sec:supp_validation}

Here,
we provide targeted supplementary validations for the main experimental claims:
training-domain robustness, the choice of LoRA over full fine-tuning, reward
causality, transformation and ghosting
design choices, and training efficiency.

\subsection{Training Domain and COCO Split Separation}

Table~\ref{tab:domain_lora_controls} reports Qwen3-VL-8B results over
three seeds. VISE is trained either on COCO or on Objects365 images, always
without captions, boxes, or category labels. The COCO training images are taken
from train2014/train2017 and evaluated on disjoint validation splits with zero
image overlap. Training on Objects365 gives nearly identical gains, showing that
the improvements are not due to COCO-specific image exposure.

\begin{table}[t]
\centering
\caption{\textbf{Training-domain and tuning-strategy validation on Qwen3-VL-8B-Instruct}
(mean$\pm$std over 3 seeds). VISE obtains consistent gains when trained on either
COCO or Objects365 images. LoRA also outperforms full fine-tuning (FFT), supporting
our frozen-encoder training design.}
\label{tab:domain_lora_controls}
\small
\setlength{\tabcolsep}{4pt}
\begin{tabular}{llccccc}
\toprule
\multirow{2}{*}{\textbf{Category}} & \multirow{2}{*}{\textbf{Training setup}}
& \multicolumn{2}{c}{\textbf{Captioning}}
& \textbf{Hallucination}
& \textbf{VQA}
& \textbf{Reasoning} \\
\cmidrule(lr){3-4} \cmidrule(lr){5-5} \cmidrule(lr){6-6} \cmidrule(lr){7-7}
& & COCO & Flickr30k & POPE\textsubscript{acc} & CaptionQA & ScienceQA \\
\midrule
Baseline
& Qwen3-VL-8B-Instruct
& $29.01{\pm}0.71$ & $34.02{\pm}0.48$
& $89.91{\pm}0.58$
& $85.21{\pm}0.29$
& $90.88{\pm}0.61$ \\
\midrule
\multirow{2}{*}{FFT}
& COCO Training
& $32.80{\pm}0.45$ & $35.86{\pm}0.61$
& $90.07{\pm}0.43$
& $85.47{\pm}0.58$
& $91.45{\pm}0.82$ \\
& Objects365 Training
& $33.51{\pm}0.50$ & $35.74{\pm}0.51$
& $90.19{\pm}0.46$
& $85.64{\pm}0.44$
& $91.48{\pm}0.55$ \\
\midrule
\multirow{2}{*}{LoRA}
& COCO Training
& $\mathbf{38.49{\pm}0.67}$ & $\mathbf{38.62{\pm}0.52}$
& $90.32{\pm}0.45$
& $86.35{\pm}0.28$
& $92.81{\pm}0.66$ \\
& Objects365 Training
& $\mathbf{38.57{\pm}0.36}$ & $\mathbf{38.71{\pm}0.46}$
& $\mathbf{90.34{\pm}0.68}$
& $\mathbf{86.52{\pm}0.67}$
& $\mathbf{92.93{\pm}0.28}$ \\
\bottomrule
\end{tabular}%
\end{table}

\subsection{LoRA versus Full Fine-Tuning}

Table~\ref{tab:domain_lora_controls} also compares full fine-tuning (FFT)
with LoRA under the same training domains. FFT improves over the base model but
consistently underperforms LoRA across captioning, hallucination, VQA, and
reasoning metrics. This supports our training design: noisy unsupervised encoder
updates can hurt visual conditioning, whereas updating the cross-modal and
decoder components is sufficient and more stable.

\subsection{Reward Causality via Random-Reward Control}

\begin{wraptable}{r}{0.55\textwidth}
\centering
\vspace{-5pt}
\caption{\textbf{Random-reward control for reward causality.} Replacing VISE's
reward with $\mathcal{R}\sim\mathcal{U}(0,1)$ leaves captioning near base
performance, while VISE gives large gains under the same training setup.}
\label{tab:random_reward_control}
\small
\setlength{\tabcolsep}{5pt}
\begin{tabular}{llccc}
\toprule
\textbf{Scale} & \textbf{Method} & COCO & NoCaps & Flickr30k \\
\midrule
\multirow{3}{*}{2B}
& Base & 21.54 & 19.52 & 26.09 \\
& Random reward & 21.38 & 19.41 & 25.87 \\
& VISE & \textbf{38.39} & \textbf{34.25} & \textbf{42.64} \\
\midrule
\multirow{3}{*}{8B}
& Base & 29.01 & 24.46 & 34.02 \\
& Random reward & 29.12 & 24.31 & 33.84 \\
& VISE & \textbf{38.49} & \textbf{34.98} & \textbf{38.62} \\
\bottomrule
\end{tabular}
\vspace{-6pt}
\end{wraptable}

To isolate the role of the invariance rewards from generic exposure to
unlabeled images, we train a random-reward control using the same images,
optimization setup, and policy update, but replace the VISE reward with
$\mathcal{R}\sim\mathcal{U}(0,1)$. Table~\ref{tab:random_reward_control} shows
that random rewards stay near base captioning performance, while VISE
produces large gains. This confirms that the improvements are driven by the
geometric and semantic invariance signals rather than by fine-tuning alone.

\subsection{Transformation and Ghosting Design Choices}

Table~\ref{tab:transform_ghosting_controls} evaluates design choices in the two
invariance branches on Qwen3-VL-2B. Moderate transformations such as affine,
crop, and flip all improve over the base model, while overly large affine
perturbations degrade performance due to degenerate border boxes. For semantic
perturbation, Gaussian ghosting performs best: the default $\sigma=25$ kernel
induces the intended visible-to-not-visible change, while weaker blur, stronger
blur, zero masking, and Gaussian noise underperform.

\begin{table}[h]
\centering
\caption{\textbf{Transformation and perturbation design validation on Qwen3-VL-2B.}
Left: geometric transformation ablations measured by COCO CIDEr.
Right: semantic perturbation ablations measured by COCO CIDEr and POPE accuracy.}
\label{tab:transform_ghosting_controls}
\small
\setlength{\tabcolsep}{5pt}
\begin{tabular}{lcc|lcc}
\toprule
\multicolumn{3}{c|}{\textbf{Geometric transformation}} &
\multicolumn{3}{c}{\textbf{Semantic perturbation}} \\
\cmidrule(lr){1-3} \cmidrule(lr){4-6}
\textbf{Config} & \textbf{COCO} & $\Delta$ &
\textbf{Config} & \textbf{COCO} & \textbf{POPE\textsubscript{acc}} \\
\midrule
Base & 21.54 & -- &
Default ghosting, $\sigma=25$ & \textbf{38.39} & \textbf{90.03} \\
Affine only & 36.84 & $+15.30$ &
Ghosting, $\sigma=50$ & 37.12 & 89.78 \\
Crop only & 36.14 & $+14.60$ &
Ghosting, $\sigma=80$ & 35.24 & 89.34 \\
Flip only & 33.84 & $+12.30$ &
Ghosting, $\sigma=15$ & 34.82 & 89.52 \\
Small affine, $\pm5^\circ$ & 35.62 & $+14.08$ &
Zero masking & 33.41 & 89.14 \\
Large affine, $\pm20^\circ$ & 31.84 & $+10.30$ &
Gaussian noise & 30.41 & 88.94 \\
\bottomrule
\end{tabular}%
\end{table}

Noisy localization steps are naturally down-weighted by the reward design:
failed localization gives near-zero REINFORCE advantage
$A_t=R_t-b_t\approx0$, while inconsistent visibility judgments yield
$\mathcal{R}_\text{sem}=0$. Thus, imperfect self-generated boxes do not provide
a strong positive learning signal.

\subsection{Training Efficiency}

Each VISE step uses seven forward passes: query generation, box prediction on
the original and transformed views, visibility prediction on the original and
ghosted views, and policy/reference log-probability evaluation. Training
Qwen3-VL-2B for 4000 steps takes 16 hours on 8$\times$ AMD MI250X GPUs using
bfloat16, about $\sim2\times$ faster to converge than multi-role self-evolving
baselines such as EvoLMM.

\begin{figure}[t]
    \centering
    % Row 1 — two side by side
    \begin{subfigure}[b]{0.45\linewidth}
        \centering
        \includegraphics[width=\linewidth]{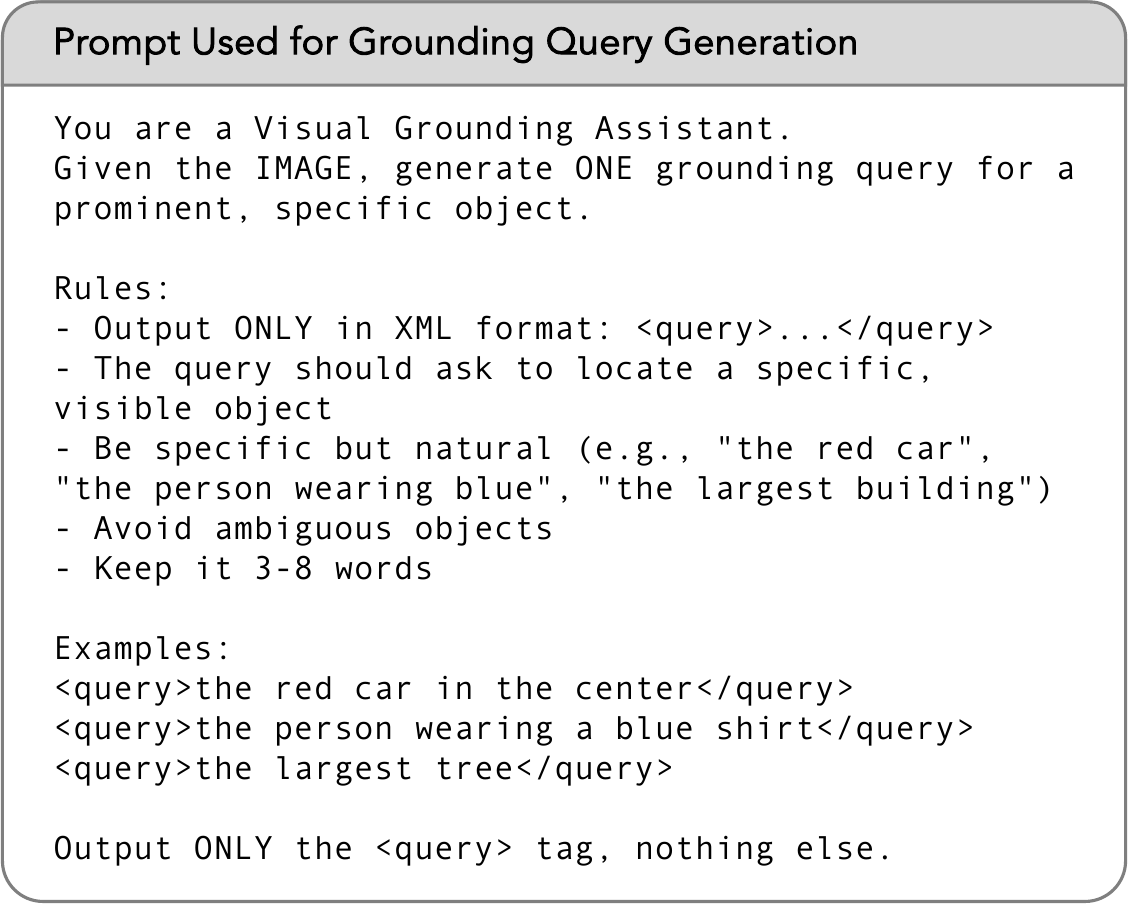}
        \label{fig:prompt1}
    \end{subfigure}
    \hfill
    \begin{subfigure}[b]{0.5\linewidth}
        \centering
        \includegraphics[width=\linewidth]{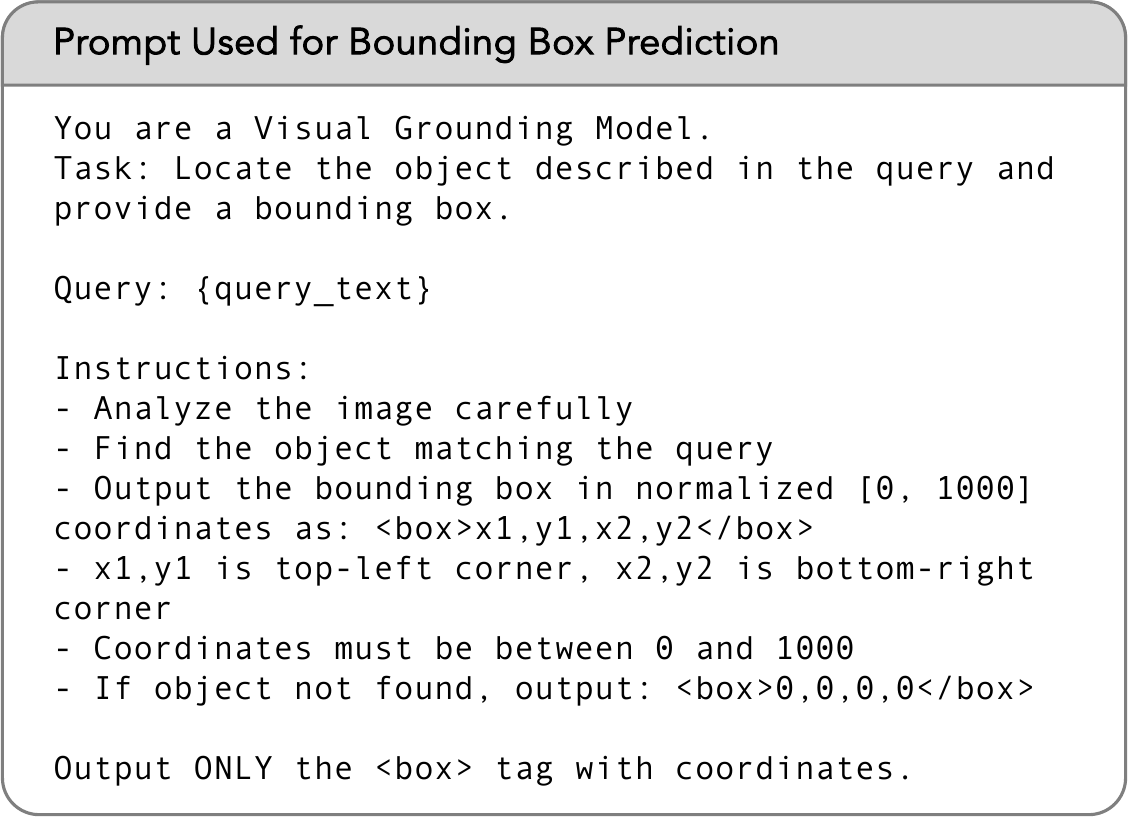}
        \label{fig:prompt2}
    \end{subfigure}
    \\[0.8em]
    % Row 2 — third centred
    \begin{subfigure}[b]{0.48\linewidth}
        \centering
        \includegraphics[width=\linewidth]{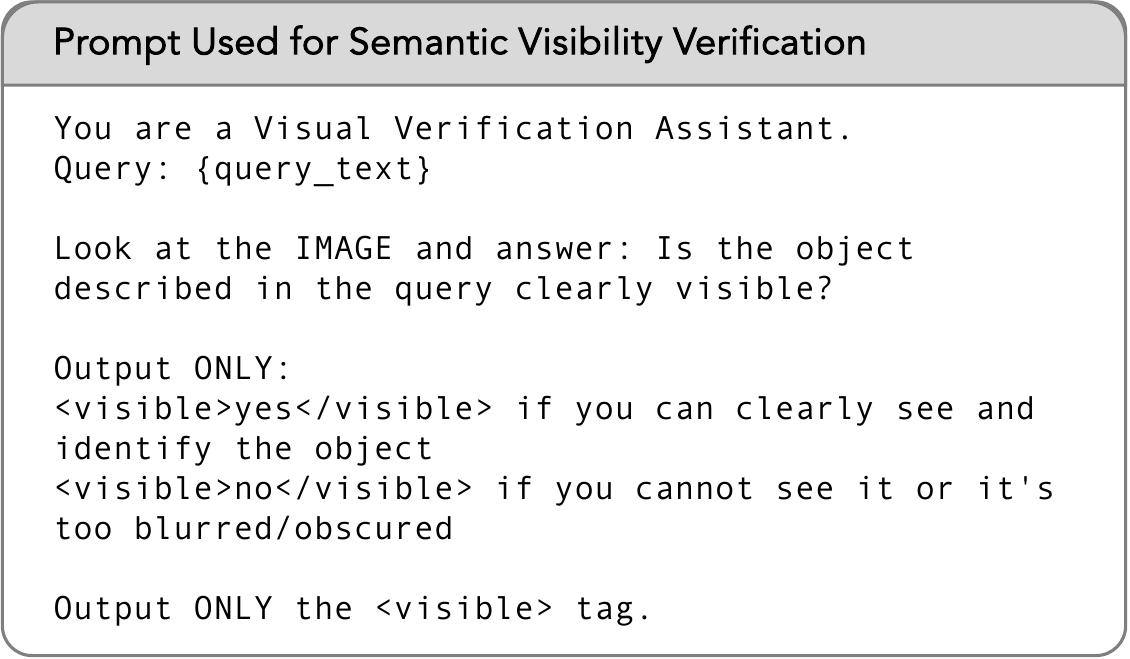}
        \label{fig:prompt3}
    \end{subfigure}
    \caption{Prompts used at each stage of VISE training.
             \textbf{(a)} Grounding query generation.
             \textbf{(b)} Bounding box prediction.
             \textbf{(c)} Semantic visibility verification.}
    \label{fig:prompts_all}
\end{figure}

\section{Prompts Used in VISE Training}

VISE uses three distinct prompts at each training step, one for 
each stage of the self-supervised pipeline. We describe each below 
and provide the corresponding prompt in the figures that follow.

\subsection{Prompt Used for Grounding Query Generation}

At the start of each training step, the model is prompted to 
generate a natural-language grounding query $q$ for the input 
image $x$. The prompt instructs the model to identify a single 
prominent, spatially unambiguous object in the scene and describe 
it concisely. The generated query is then used as input for both 
the bounding box prediction and the semantic visibility verification 
stages within the same step. No external query bank, template, or 
category list is used --- the query is produced entirely by the 
model itself from the raw image.

\subsection{Prompt Used for Bounding Box Prediction}

Given the image $x$ (or its geometrically transformed version 
$x' = \mathcal{T}(x)$) and the generated query $q$, the model 
is prompted to predict a bounding box localizing the queried 
object. The prompt specifies the normalized coordinate space 
$[0, 1000]^4$ and the expected output format. This same prompt 
structure is used for both the original and transformed views, 
ensuring that any difference in predicted boxes reflects the 
model's visual grounding behavior rather than prompt variation.

\subsection{Prompt Used for Semantic Visibility Verification}

After predicting $B_\text{orig}$, the model is prompted to 
assess whether the queried object is clearly visible in both the 
original image $x$ and the ghosted image $\tilde{x}$. The prompt 
asks for a binary yes/no judgment and is kept deliberately minimal 
to avoid leading the model toward a particular answer. The 
semantic invariance reward $\mathcal{R}_\text{sem}$ is computed 
from the pair of visibility judgments returned by this prompt, 
as described in the Method Section of the main paper.

\section{Extended Qualitative Results}

\subsection{Generation-Time Visual Attention}

\begin{figure}[t]
    \centering
    \includegraphics[width=\linewidth]{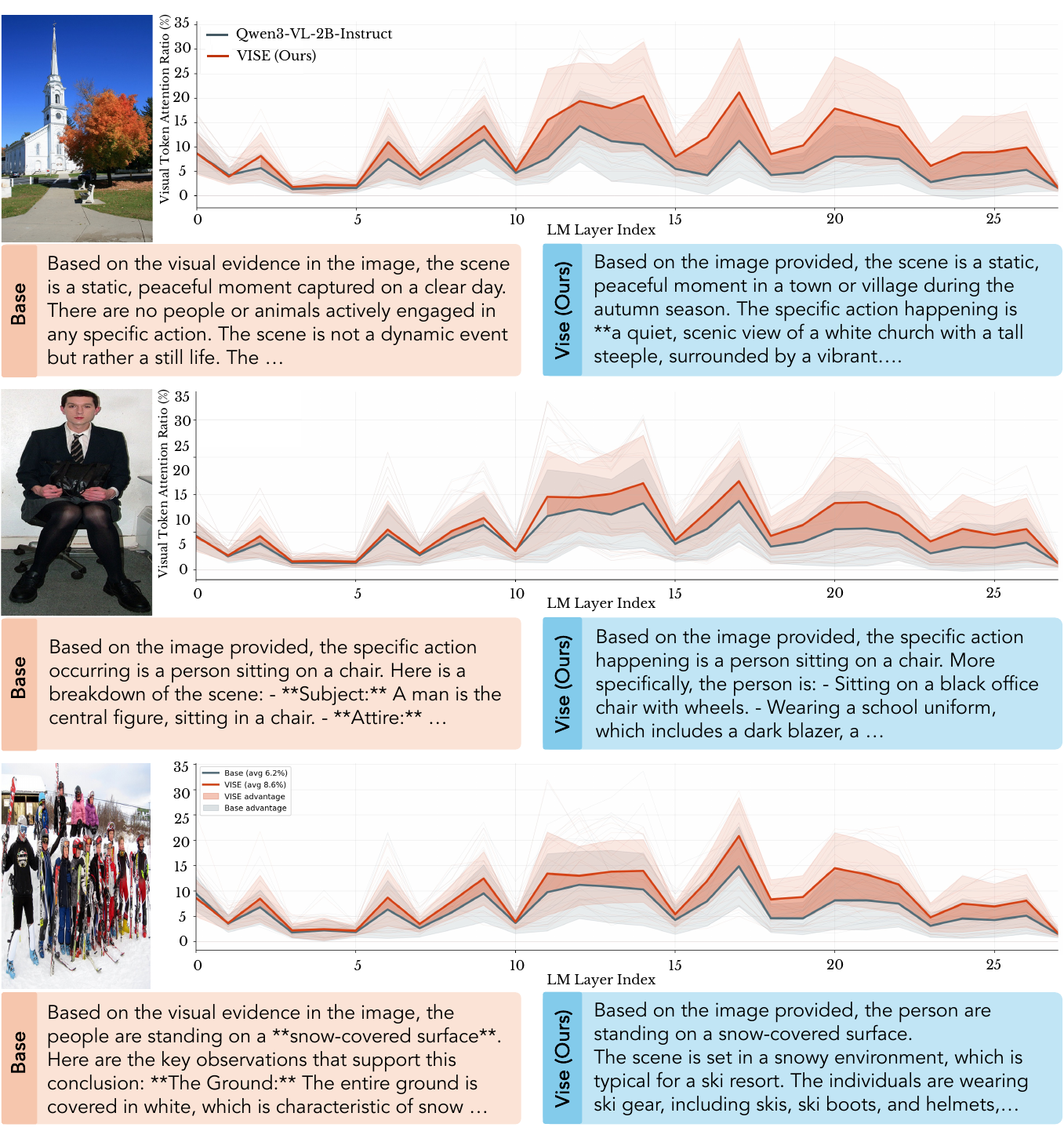}
    \caption{Additional generation-time visual attention comparisons between 
    the base model and VISE on Qwen3-VL-2B. For each example, the attention 
    plot shows the fraction of attention allocated to image tokens per decoder 
    layer, alongside the corresponding outputs from both models. VISE 
    consistently attends more to visual tokens across mid-to-late layers, 
    producing more grounded and specific descriptions.}
    \label{fig:qual_attn1}
\end{figure}

\begin{figure}[t]
    \centering
    \includegraphics[width=\linewidth]{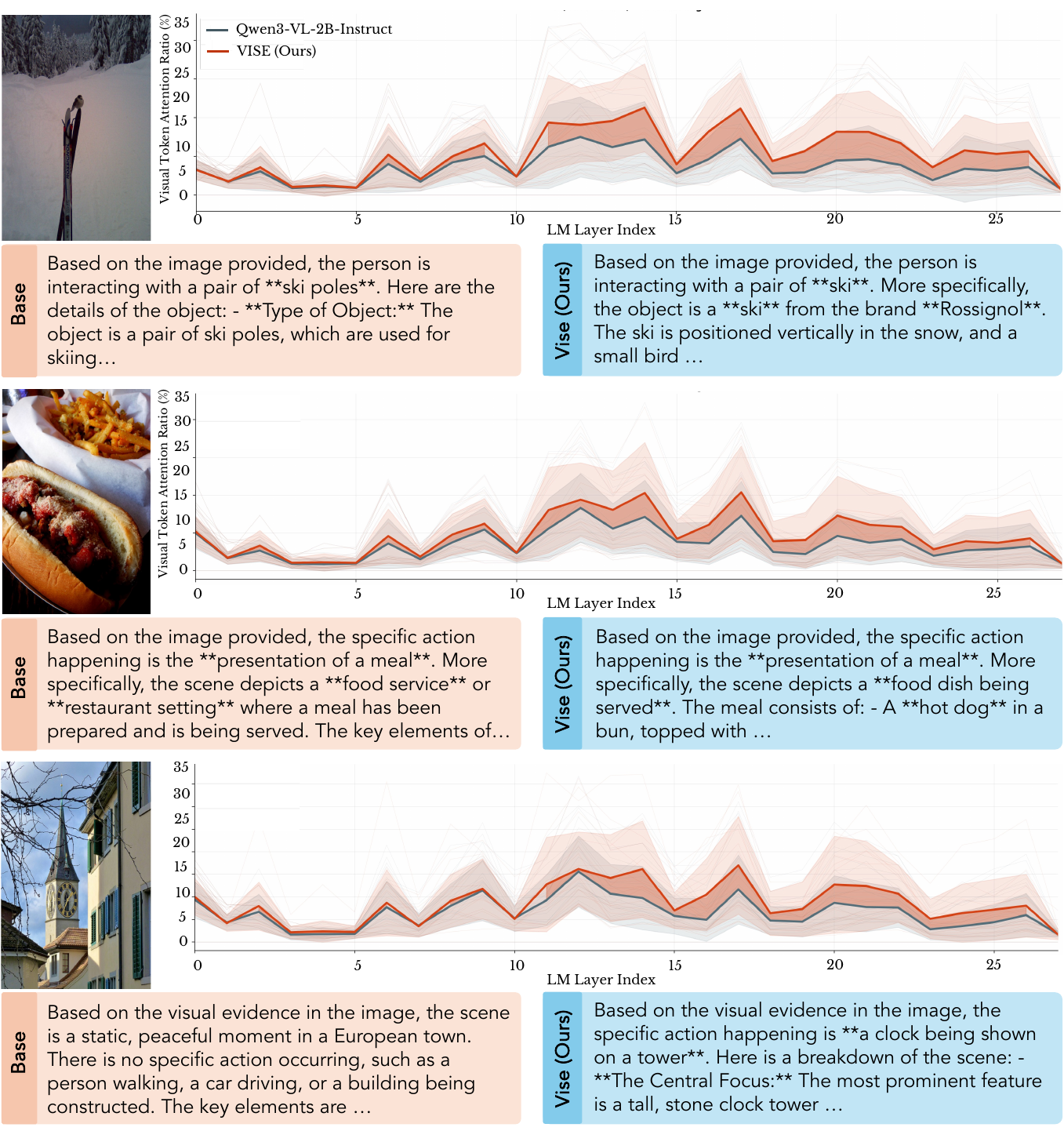}
    \caption{Additional generation-time visual attention comparisons (continued).}
    \label{fig:qual_attn2}
\end{figure}

\begin{figure}[t]
    \centering
    \includegraphics[width=\linewidth]{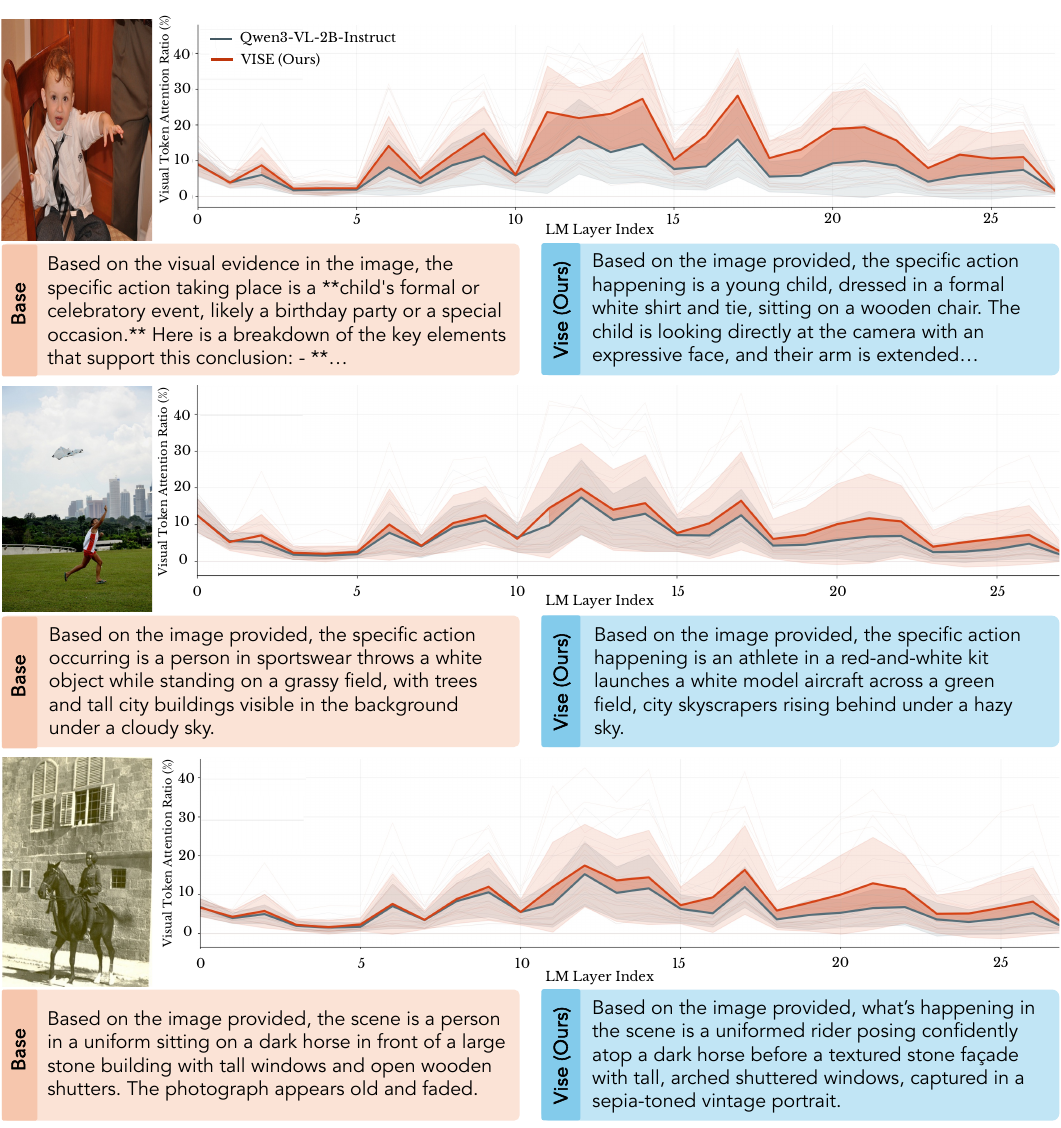}
    \caption{Additional generation-time visual attention comparisons (continued).}
    \label{fig:qual_attn3}
\end{figure}

\begin{figure}[t]
    \centering
    \includegraphics[width=\linewidth]{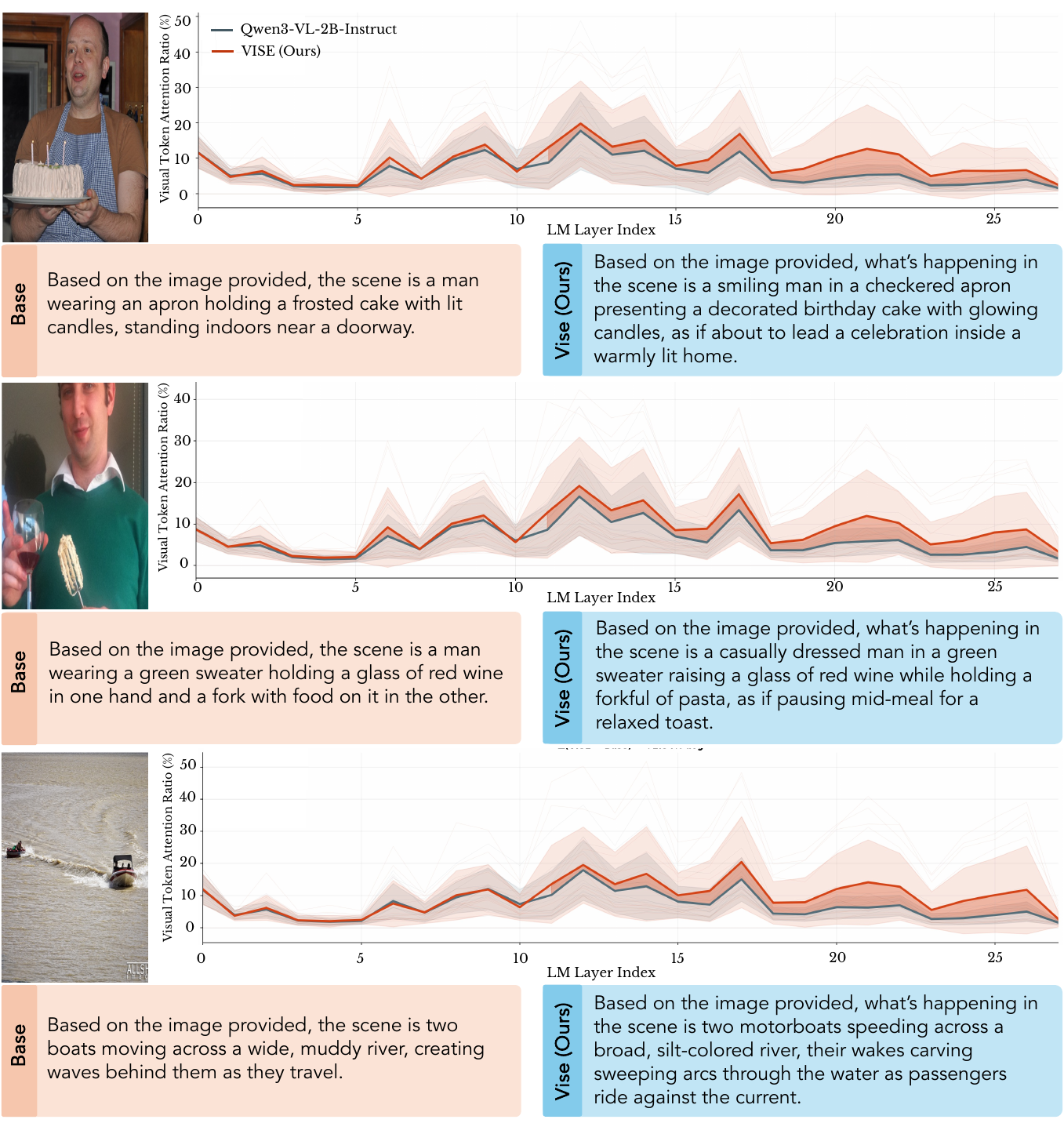}
    \caption{Additional generation-time visual attention comparisons (continued).}
    \label{fig:qual_attn4}
\end{figure}

\begin{figure}[t]
    \centering
    \includegraphics[width=\linewidth]{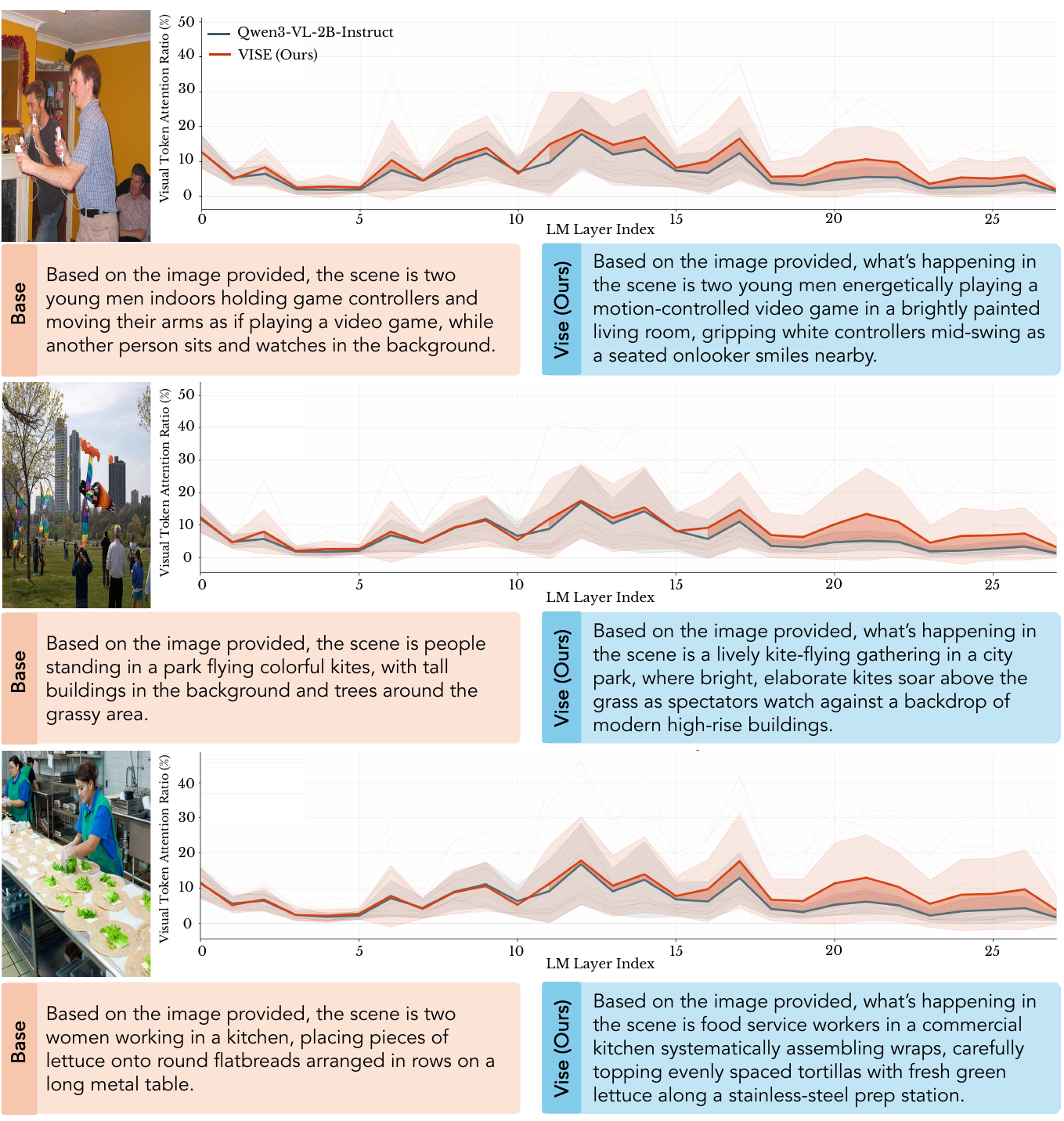}
    \caption{Additional generation-time visual attention comparisons (continued).}
    \label{fig:qual_attn5}
\end{figure}

Figures~\ref{fig:qual_attn1}–\ref{fig:qual_attn5} show additional per-sample, generation-time visual attention comparisons between the base Qwen3-VL-2B model~\cite{bai2025qwen3vltechnicalreport} and VISE across a range of scene types using the prompt, "What is happening in this scene?" For each example, we plot the fraction 
of attention allocated to image tokens at each decoder layer 
during generation, alongside the corresponding text outputs from 
both models. The attention advantage of VISE is consistent across 
all examples shown, concentrating in the mid-to-late decoder 
layers where semantic generation decisions are made, and is 
accompanied by noticeably more specific and visually grounded 
output text.

\subsection{Image Description Comparisons}

\begin{figure}[t]
    \centering
    \includegraphics[width=\linewidth]{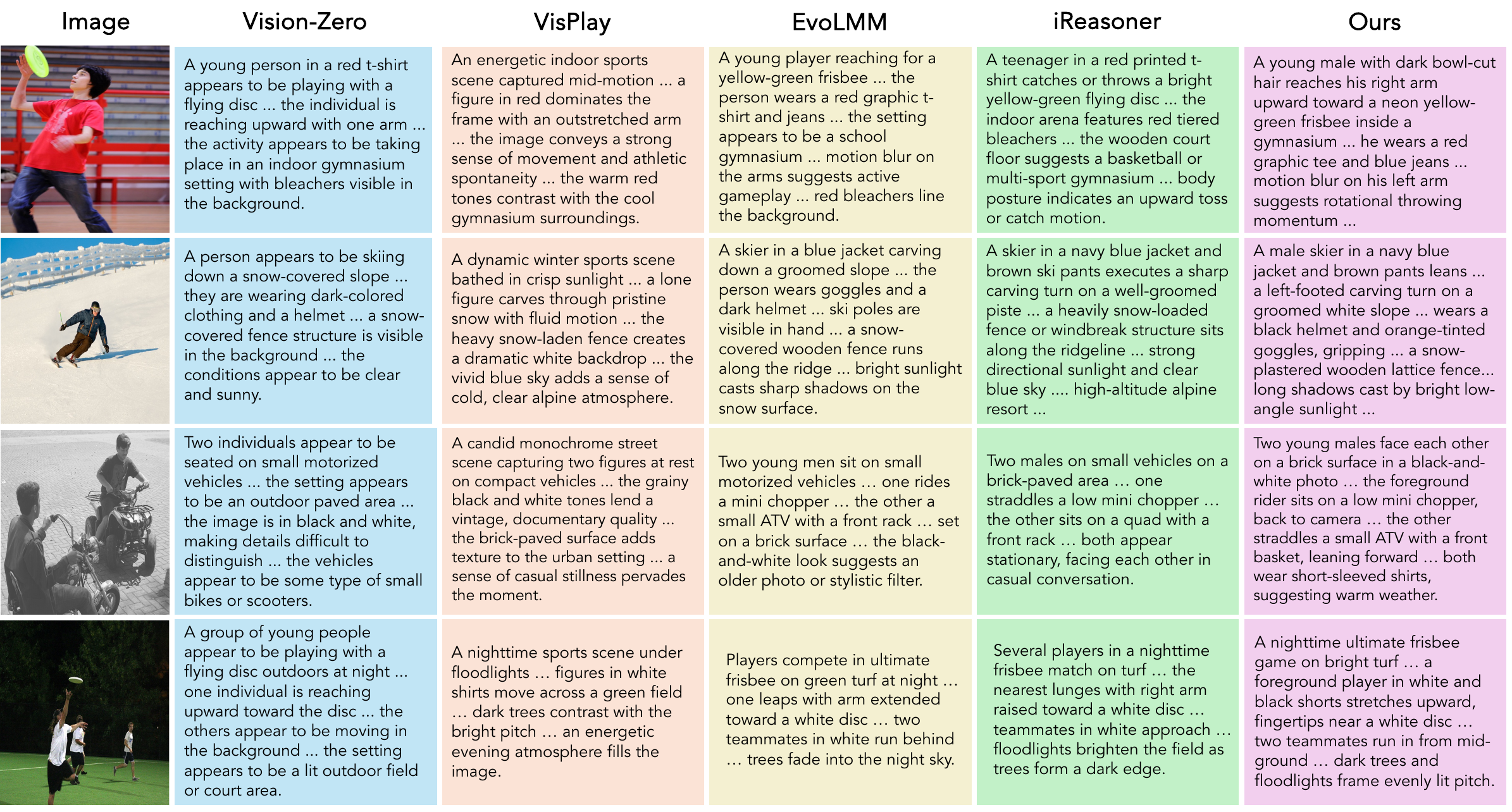}
    \caption{Additional qualitative comparisons of VISE against all baselines 
    across diverse scene types. VISE consistently produces more specific and 
    visually grounded descriptions, correctly identifying fine-grained details 
    such as clothing, object types, and spatial relationships that baselines 
    either miss or describe only at a category level.}
    \label{fig:qual_desc1}
\end{figure}

\begin{figure}[t]
    \centering
    \includegraphics[width=\linewidth]{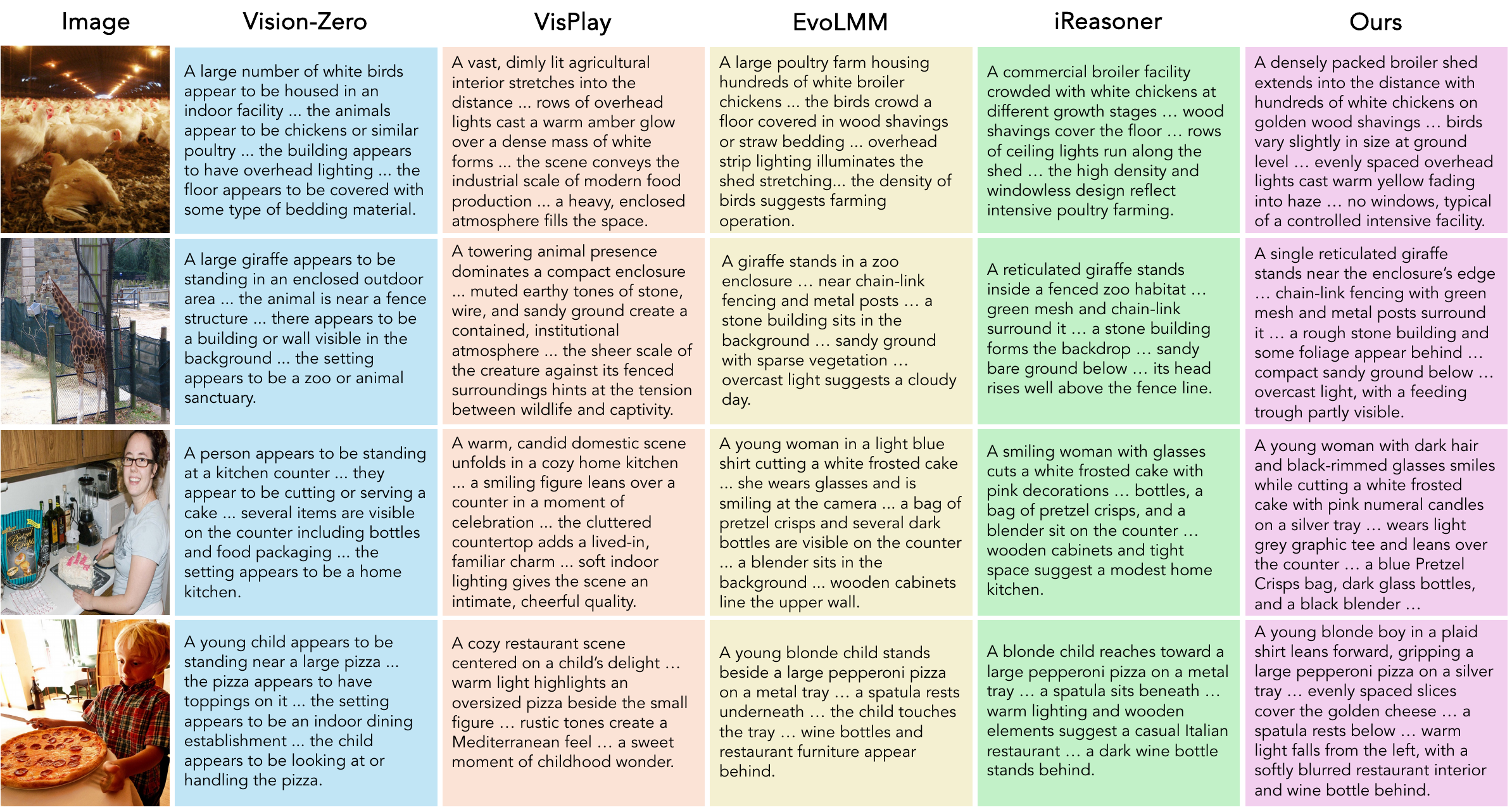}
    \caption{Additional qualitative comparisons (continued).}
    \label{fig:qual_desc2}
\end{figure}

\begin{figure}[t]
    \centering
    \includegraphics[width=\linewidth]{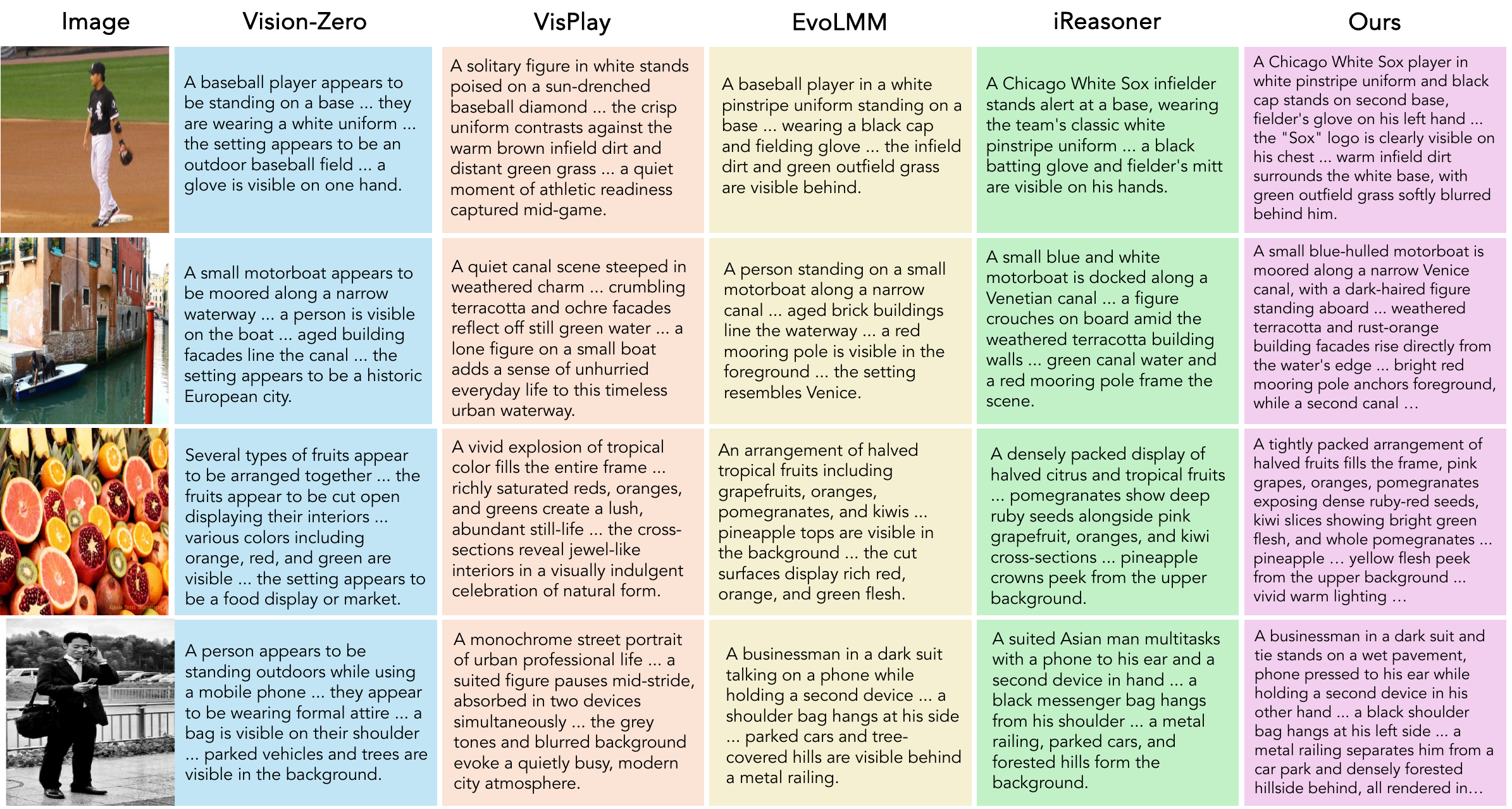}
    \caption{Additional qualitative comparisons (continued).}
    \label{fig:qual_desc3}
\end{figure}

\begin{figure}[t]
    \centering
    \includegraphics[width=\linewidth]{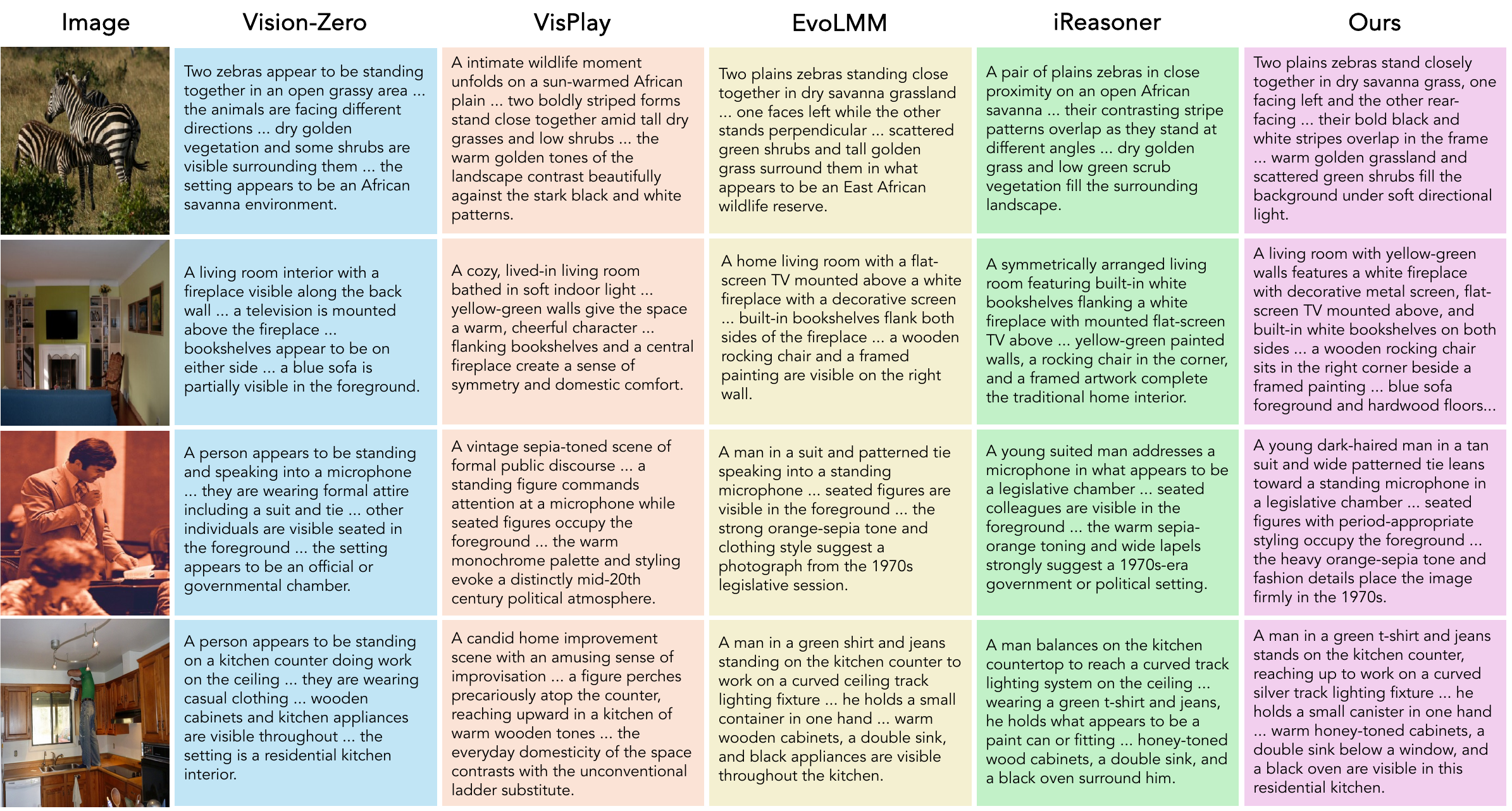}
    \caption{Additional qualitative comparisons (continued).}
    \label{fig:qual_desc4}
\end{figure}

Figures~\ref{fig:qual_desc1} to \ref{fig:qual_desc4} provide 
additional qualitative comparisons of image descriptions produced 
by VISE and four self-evolving baselines: VisionZero (RealWorld)~\cite{visionzero}, 
VisPlay~\cite{p7}, EvoLMM~\cite{p3}, and iReasoner~\cite{p1}. All methods are evaluated on the 
same images under the same prompts using their respective 
Qwen3-VL-2B checkpoints. Baseline methods tend to produce 
descriptions that are either vague and category-level, relying 
on statistically common scene descriptions, or occasionally 
confident but incorrect about specific visual details. VISE 
consistently produces more fine-grained and accurate descriptions, 
correctly identifying object-level details such as clothing 
attributes, vehicle types, spatial relationships, and scene-specific 
context that the baselines miss or misattribute.

% Attention figure 1

% Attention figure 2

% Attention figure 3

% Attention figure 4

% Attention figure 5

% Qualitative comparison figure 1

% Qualitative comparison figure 2

% Qualitative comparison figure 3

% Qualitative comparison figure 4

\subsection{Per-Sample Generation-Time Attention Breakdown}

\begin{figure}[t]
    \centering
    \includegraphics[width=\linewidth]{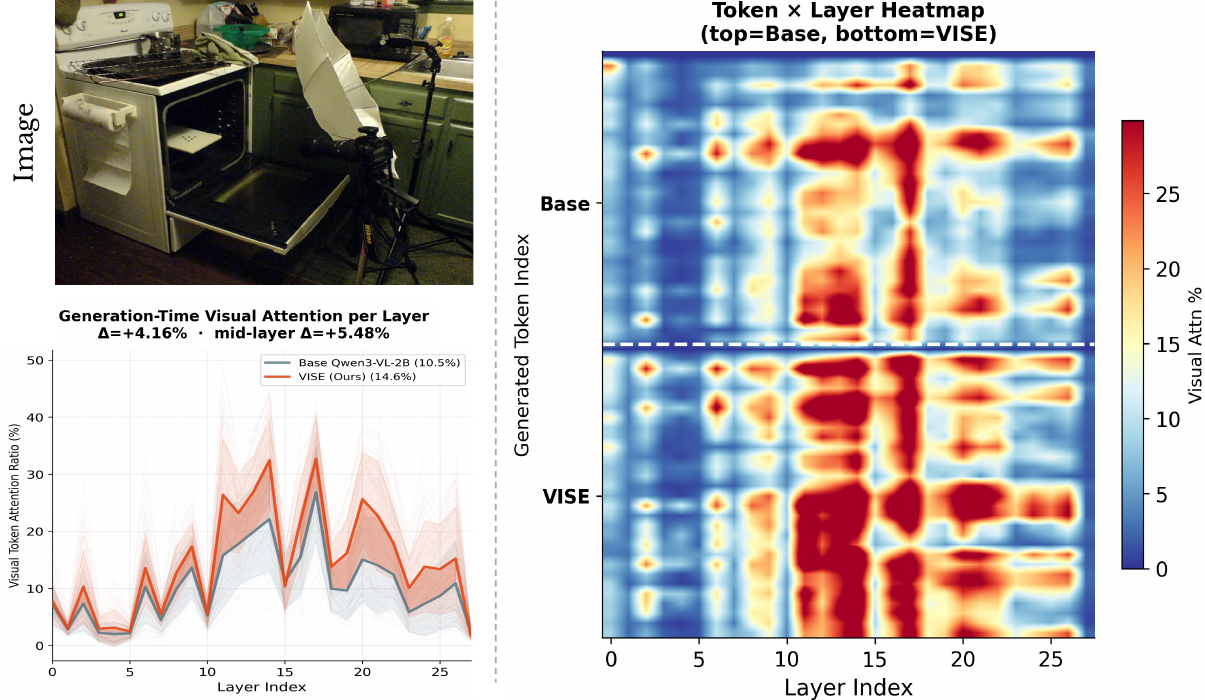}
    \caption{Per-sample generation-time attention breakdown on 
    Qwen3-VL-2B. \textbf{Left:} input image and per-layer 
    visual token attention ratio for the base model and VISE. 
    \textbf{Right:} Token$\times$Layer heatmap with base model 
    (top) and VISE (bottom); red regions indicate high visual 
    attention.}
    \label{fig:attn_2b_r1}
\end{figure}

\begin{figure}[t]
    \centering
    \includegraphics[width=\linewidth]{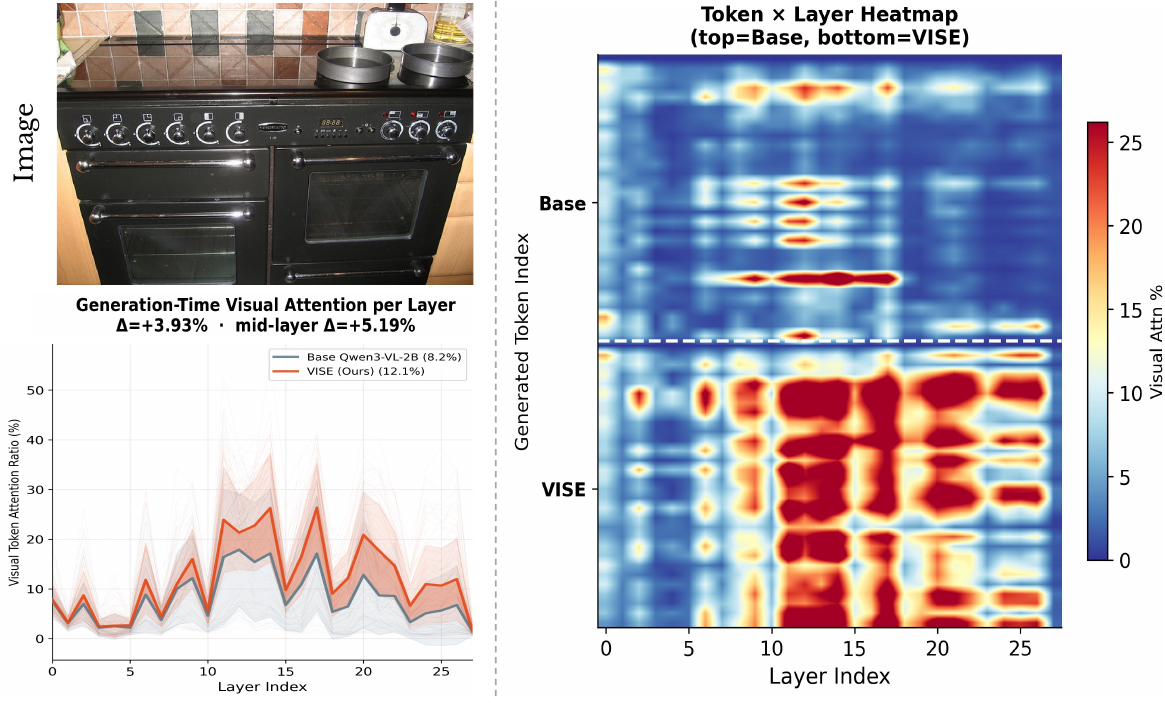}
    \caption{Per-sample generation-time attention breakdown on 
    Qwen3-VL-2B (sample 2).}
    \label{fig:attn_2b_r2}
\end{figure}

\begin{figure}[t]
    \centering
    \includegraphics[width=\linewidth]{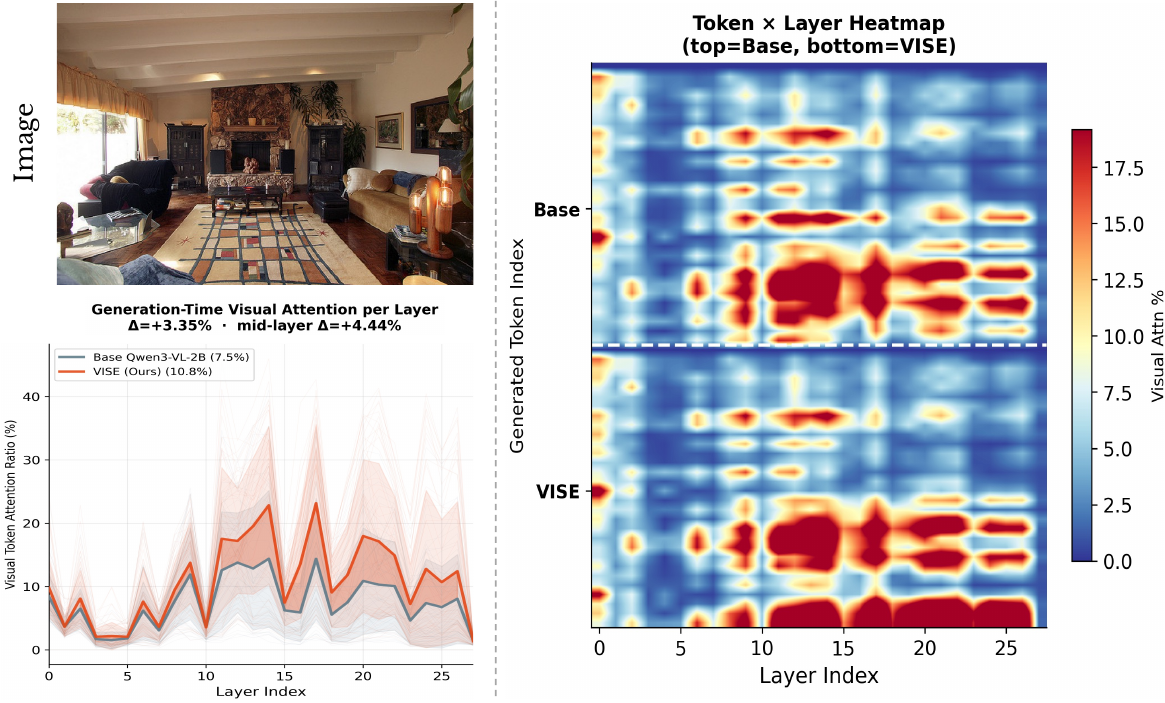}
    \caption{Per-sample generation-time attention breakdown on 
    Qwen3-VL-2B (sample 3).}
    \label{fig:attn_2b_r3}
\end{figure}

\begin{figure}[t]
    \centering
    \includegraphics[width=\linewidth]{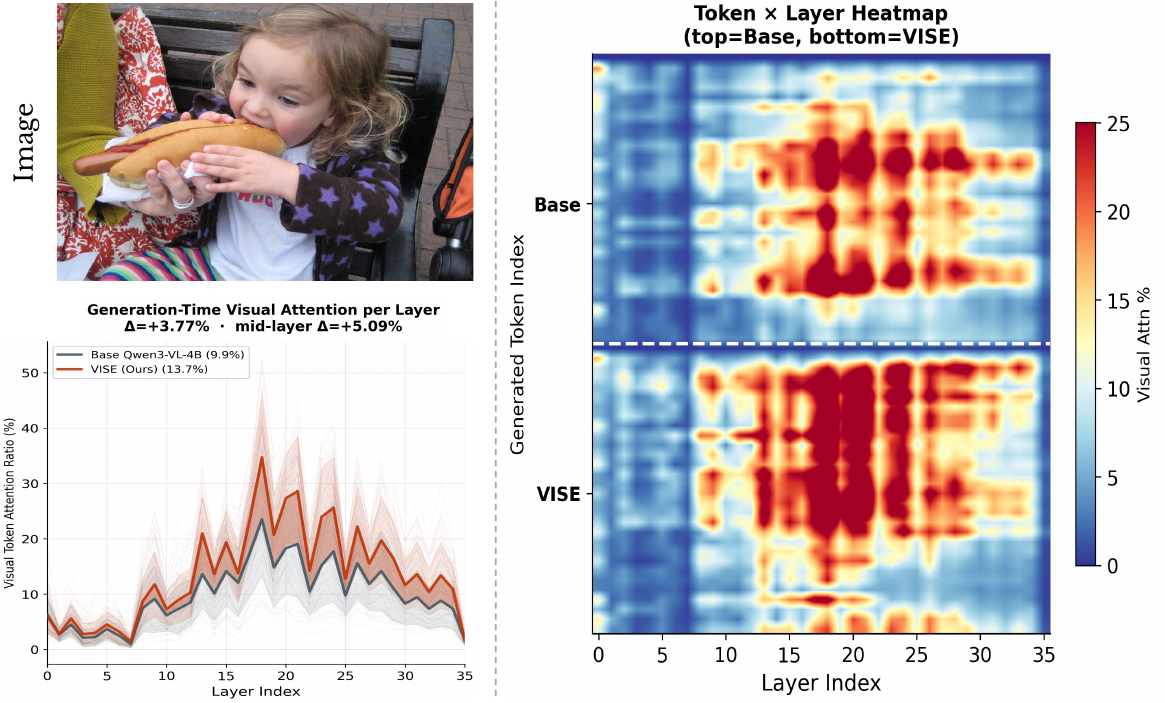}
    \caption{Per-sample generation-time attention breakdown on 
    Qwen3-VL-4B. The attention advantage concentrates in 
    layers 15--25, consistent with the aggregate result in 
    Figure~6 of the main paper.}
    \label{fig:attn_4b_r1}
\end{figure}

\begin{figure}[t]
    \centering
    \includegraphics[width=\linewidth]{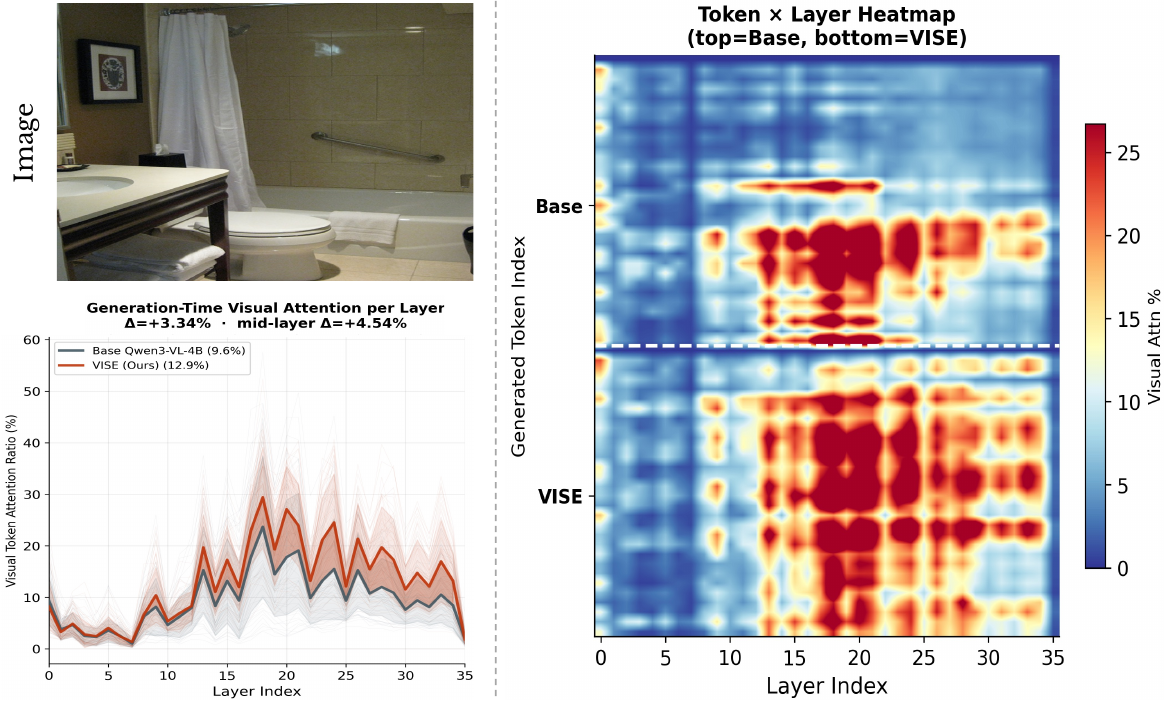}
    \caption{Per-sample generation-time attention breakdown on 
    Qwen3-VL-4B (sample 2).}
    \label{fig:attn_4b_r2}
\end{figure}

\begin{figure}[t]
    \centering
    \includegraphics[width=\linewidth]{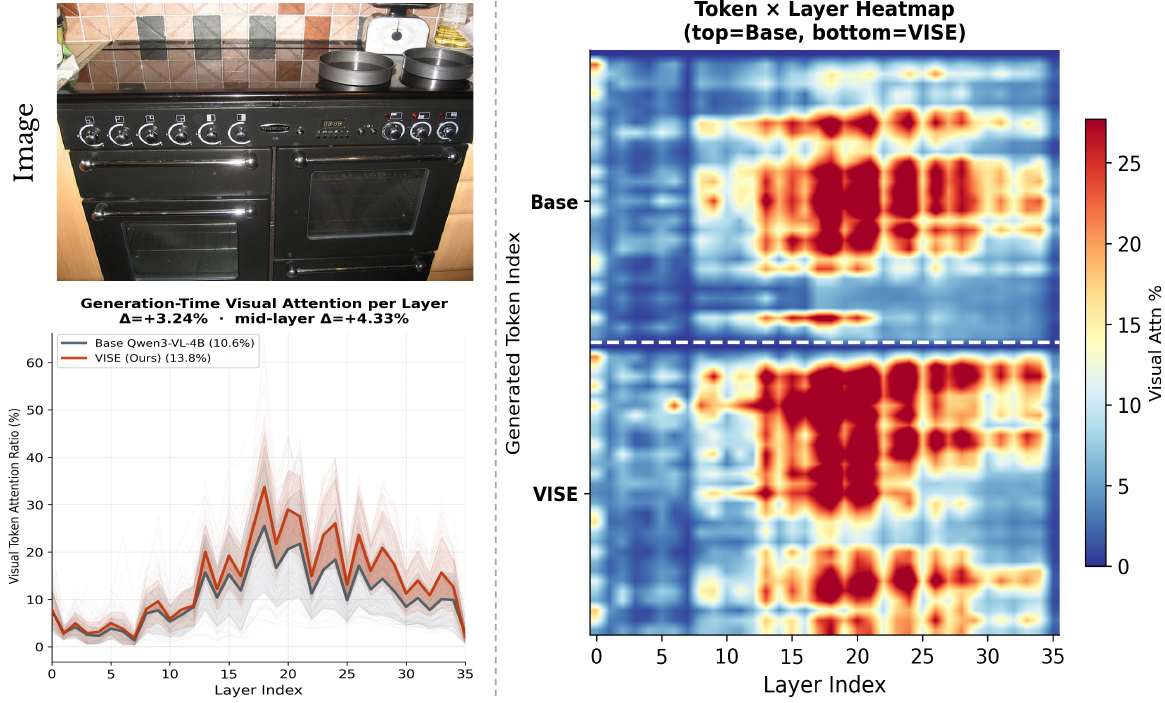}
    \caption{Per-sample generation-time attention breakdown on 
    Qwen3-VL-4B (sample 3).}
    \label{fig:attn_4b_r3}
\end{figure}

Figures~\ref{fig:attn_2b_r1}--\ref{fig:attn_2b_r3} and 
Figures~\ref{fig:attn_4b_r1}--\ref{fig:attn_4b_r3} show 
per-sample generation-time attention breakdowns for randomly 
selected samples from the COCO dataset, evaluated on 
Qwen3-VL-2B and Qwen3-VL-4B respectively. Each figure shows 
the input image, a per-layer line plot of the visual token 
attention ratio during generation, and a Token$\times$Layer 
heatmap comparing the base model (top) and VISE (bottom) side 
by side. The line plots show that VISE consistently allocates 
a higher fraction of attention to image tokens throughout 
generation, with the advantage most pronounced in mid-to-late 
layers where semantic content is produced. The heatmaps make 
this difference concrete at the token level: VISE shows broader 
and more intense red regions across both the layer and token 
axes, indicating that individual generated tokens are more 
strongly anchored to visual evidence. This pattern holds across 
all six samples and both model scales, and is particularly 
strong in cases requiring spatial layout description and 
fine-grained visual detail. All samples are evaluated using the prompt: \textit{``Describe in 
detail the spatial layout and positions of all objects in this image.''}

% -------- 2B figures --------

% -------- 4B figures --------

